\documentclass[a4paper,fleqn]{cas-dc}

\usepackage[authoryear,longnamesfirst]{natbib}
\usepackage{subcaption}

\usepackage{tabularx, array}
\usepackage{graphicx}
\usepackage{booktabs}
\usepackage{adjustbox}
\usepackage{hyperref}

\def\tsc#1{\csdef{#1}{\textsc{\lowercase{#1}}\xspace}}
\tsc{WGM}
\tsc{QE}

\begin{document}
\let\WriteBookmarks\relax
\def\floatpagepagefraction{1}
\def\textpagefraction{.001}

\shorttitle{}    

\shortauthors{}  

\title [mode = title]{ Review and Evaluation of Point-Cloud based Leaf Surface Reconstruction Methods for Agricultural Applications}  


\tnotetext[1]{} 

%

\author[1]{Arif Ahmed}


\ead{arifa@unr.edu}


\credit{Methodology, Software, Investigation,  Visualization, Writing - Original Draft}

\author[1]{Parikshit Maini}


\ead{pmaini@unr.edu}
\cormark[1]


\credit{Conceptualization, Methodology, Writing - Review \& Editing, Supervision, Funding acquisition}

\affiliation[1]{organization={Department of Computer Science and Engineering, University of Nevada, Reno},\\
            city={\\ Reno},
            postcode={89557}, 
            state={Nevada},
            country={USA}}







\cortext[1]{Corresponding author}


\nonumnote{This work was funded by University of Nevada, Reno.}

\begin{abstract}
Accurate reconstruction of leaf surfaces from 3D point cloud is essential for agricultural applications such as  phenotyping. 
However, real-world plant data (i.e., irregular 3D point cloud) are often complex to reconstruct plant parts accurately.
A wide range of surface reconstruction methods has been proposed, including parametric, triangulation-based, implicit, and learning based approaches, yet their relative performance for leaf surface reconstruction remains insufficiently understood. 

In this work, we present a comparative study of nine representative surface reconstruction methods for leaf surfaces. We evaluate these methods on three publicly available datasets: LAST-STRAW, Pheno4D, and Crops3D - spanning diverse species, sensors, and sensing environments, ranging from clean high-resolution indoor scans to noisy low-resolution field settings. The analysis highlights the trade-offs between surface area estimation accuracy, smoothness, robustness to noise and missing data, and computational cost across different methods.
These factors affect  the cost and constraints of robotic hardware used in agricultural applications. 

Our results show that each method exhibits distinct advantages depending on application and resource constraints. The findings provide practical guidance for selecting surface reconstruction techniques for resource constrained robotic platforms. 
\end{abstract}




\maketitle

\section{Introduction}
\label{sec:1_intro}
The high labor and costs associated with manual plant phenotyping  \cite{stausberg20243d} have motivated the development of automated solutions in agricultural robotics \cite{ahmed2025saral}. These approaches perform plant phenotyping by analyzing plant geometry reconstructed from 3D point cloud data \cite{stausberg20243d}. These data are typically acquired through either active sensing methods, such as LiDAR, or passive approaches using camera-based techniques such as Structure-from-Motion (SfM) \cite{santos20143d,stausberg20243d}. Leaf surface reconstruction from 3D point cloud data is a core problem in plant phenotyping because it describes plant health, growth, and yield estimation \cite{ando2021robust}.
Beyond structural analysis, accurate 3D leaf surface reconstruction also helps scientists study how leaves perform photosynthesis, exchange gases, and transport nutrients \cite{loch2005application,hui2026leaflods}.

From phenotyping perspective, reconstructed leaf surfaces allow for the extraction of geometric traits such as leaf length and width.
In contrast, plant scientists focus more on modeling a leaf surface for applications such as modeling agrichemical spray droplet movement and spread on surface \cite{kempthorne2015surface, mayo2015simulating}. In this context,  leaf surface reconstruction is critical, as variations in surface characteristics strongly influence computational predictions of droplet interception, impaction behavior (adhere, bounce, or shatter), and overall spray retention \cite{dorr2014towards}. 
Furthermore, accurate reconstruction of the leaf surface is important for critical phenotypic traits such as leaf size, leaf area index, and leaf angle distribution \cite{zermas20203d,stausberg20243d, li2025applications}.

Despite the wide range of available surface reconstruction methods, their suitability for reconstruction leaf surfaces in real-world agricultural robotics applications remains insufficiently understood. In particular, there is a lack of systematic evaluation on real plant dataset that quantifies trade-offs between surface are estimation accuracy, qualitative surface fidelity, and computational efficiency under constraints imposed by robotic platforms  and onboard computation.

This survey presents a comprehensive review of  leaf surface reconstruction methods from the perspective of  agricultural robotics. 
We analyze representative approaches from each methodological category and conduct a comparative study on real leaf point cloud datasets. 
Furthermore, we examine  the strength and limitations of each method, providing practical insight to guide the selection of appropriate techniques for different robotic platforms and onboard computational resources.

\newpage
\section{Literature Review}
Prior works classified leaf surface reconstruction methods into two broad categories: model-free (bottom-up) and model-based (top-down) \cite{stausberg20243d}. The model-free methods do not require any parameter calibration or knowledge about the leaf geometry. These methods rely on local information of the point cloud data to reconstruct leaf surfaces such as Poisson surface reconstruction, Ball-Pivoting-Algorithm (BPA), non-uniform rational B-spline (NURBS) \cite{wang2013geometric, stausberg20243d}.
The model-based methods utilize prior knowledge to generate a mathematical model of the surface \cite{quan2006image}. They can better capture the morphological differences but can not capture subtle differences within each leaf \cite{ando2021robust}. However, model-free methods suffer from noisy or missing data as they do not exploit the prior knowledge. 
However,  \cite{lim2014surface} reported different classifications for surface reconstruction from the literature in their review. They reviewed two representations: explicit and implicit \cite{zhao2001fast}. 
The explicit surfaces denote the exact location of a surface, popular representations are parametric (e.g., B-Spline)  and triangulated surface (e.g., Delaunay) \cite{piegl2012nurbs, dimitrov2016non} . 
The parameterization is a challenge for 3D surface reconstruction and noisy data can make it more difficult. For triangulation, correct connectivity for the sample points can be difficult for non-uniform and noisy data. 
On the other hand, implicit surfaces (a.k.a volumetric representations) represent a surface as a particular isocontour of a scalar function (e.g., Poisson).
Another popular surface reconstruction is 
Kohonen network or Self-Organizing Map (SOM) \cite{kohonen1997exploration}, \cite{lim2014surface} classified it as a neural network method. 
This iterative method can better capture the surface topology. 
This  method is useful when leaf surface reconstruction has missing leaf points or holes.

\section{Organization}
The remainder of this paper is organized as follows. 
In Section~\ref{sec:all_lsr_methods}, we present nine representative leaf surface reconstruction methods from the literature and their mathematical formulation. 
In Section~\ref{sec:3_performance}, we evaluate these leaf surface reconstruction methods experimentally, covering dataset preparation, implementation details, and reconstruction results. 
In  Section~\ref{sec:4_robotic_percept}, we discuss robotic perception challenges in agricultural environments and real-time deployment constraints.  
In Section~\ref{sec:5_open_challenge}, we identify open challenges and research directions, and we conclude   in Section \ref{sec:6_conclusion}.

\section{Leaf Surface Reconstruction Methods}
\label{sec:all_lsr_methods}
In this section, we describe nine leaf surface reconstruction methods and their mathematical formulation. These methods are: 
\begin{enumerate}
    \item  B-Spline
    \item Non-Uniform Rational B-Spline (NURBS) 
    \item Discrete Smoothing $D^2$-Spline (D2S) 
    \item Delaunay  
    \item  Moving Least Squares (MLS) 
    \item Locally estimated scatterplot smoothing
(LOESS)
    \item Poisson Surface Reconstruction 
    \item  Ball Pivoting Algorithm (BPA) 
    \item Self-organizing Map (SOM). 
\end{enumerate}

\begin{table*}
\caption{Comparison: leaf surface reconstruction methods}\label{tbl1}
\begin{tabular*}{\tblwidth}{@{}lccccCc@{}}
 
\toprule
  \textbf{Method} & \textbf{Category}   & \textbf{Input}  & \makecell{\textbf{Noise}\\ \textbf{Robustness} }&  \makecell{\textbf{Missing} \\ \textbf{Data}}& \textbf{Surface} & \textbf{Key Parameters} \\
\midrule
B-Spline & \makecell{Spline-based,\\ parametric} & grid data & Low & Partial &  \makecell{Parametric \\surface} & Control point net\\
\midrule
NURBS & \makecell{Spline-based, \\parametric rational} & grid data & Medium & Partial & \makecell{Rational \\ surface} & \makecell{Better than B-spline \\ for rational weights }\\
\midrule
 $D^2$-Spline & \makecell{Spline-based, \\ variational,\\ bending energy} &   point cloud & High  & Yes & \makecell{Explicit \\ surface} & \makecell{GCV auto-selects \\smoothing parameter}\\
 \midrule
 MLS & \makecell{Fitting, \\polynomial} &  point cloud (2D) & Medium & Yes &  \makecell{Explicit \\ surface} & \makecell{ neighborhood size \\ (bandwidth)} \\
 \midrule
 
 LOESS & \makecell{Fitting, \\tricube-weighted\\ local polynomial} &  \makecell{point cloud  (2D)} & High & Yes & \makecell{Explicit \\ surface} & \makecell{Local neighborhood \\ per query} \\
 \midrule
Delaunay 2.5D & Triangulation & point cloud (2D) & Low & No &\makecell{ Triangulated \\mesh }& None\\
\midrule
Poisson & Poisson, Implicit & \makecell{point cloud\\ with normals }& High & Yes & \makecell{Watertight \\mesh} & \makecell{Octree depth, \\ bumpy surface \\ for noisy leaf} \\
\midrule
BPA 
& \makecell{Ball pivoting,\\ triangulation} &\makecell{point cloud\\ with normals }& \makecell{Low } & No & \makecell{ Triangulated \\mesh } & \makecell{Sensitive to ball-radius, \\ produce holes where \\ ball cannot reach }\\
\midrule
SOM 
& \makecell{Competitive\\ learning \\ (Unsupervised)} & raw point cloud  & Medium & Yes & \makecell{Discrete \\ neuron \\ mesh }& \makecell{Grid size, epoch \\ captures leaf topology}\\
\bottomrule
\end{tabular*}
\end{table*}

\subsection{B-Spline}
A B-spline (i.e., basis-spline) surfaces represent one of the most widely used formulations for geometric surface reconstruction from point cloud data \cite{piegl2012nurbs}. 
To construct the geometric model of a broad-leaf plants leaf veins and surfaces, \cite{wang2013geometric} used this parametric method. \cite{wu2021parametric} also used this method for modeling tea leaves.   
\cite{piegl2012nurbs} define a B-spline surface by taking a bidirectional net of control points, two knot vectors, and the products of the univariate B-spline functions. So, B-spline surface of degree $p\times q$ as: 

\begin{equation} \mathbf{S}(u,v) = \sum_{i=0}^n \sum_{j=0}^m N_{i,p}(u) N_{j,q}(v) ~\mathbf{P}_{ij} \end{equation}

where $\mathbf{P}_{ij} \in \mathbb{R}^3 $ are the vertices of the control point vertices arranged in a $n\times m $ control net, and $N_{i,p}(u), ~ N_{j,q}(v)$ are the B-spline basis functions in the $u$ and $v$ directions respectively, each constructed from their own knot vector via Cox-de Boor recursion. 
The basis functions are piecewise polynomial and smooth at their joins. Because 
each basis function is nonzero over only a small range of parameter values, each control point influences only a local region of the surface rather than the entire surface. 
To fit this surface to measured data, \cite{piegl2012nurbs} assign parameter values to each data point using chord-length parameterization. Here, the data points that are far apart in physical space receive a larger gap in parameter space, and data points that are close together in physical space receive a small gap. This ensures the fitted surface does not compress or stretch unevenly across the leaf. 
Once parameter values are assigned, the method finds the control points by enforcing: \textit{ the surface must pass exactly through every data point}. This turns the fitting problem into a standard linear system, where the unknowns are the control point positions and each equation corresponds to one data point. Solving this system gives a surface that reproduces the measured geometry of the leaf. However, the surface also reproduces any noise in the point cloud data.

\subsection{Non-Uniform Rational B-Spline (NURBS)}
NURBS generalize polynomial B-splines by introducing a positive weight $w_{ij}$ for each control point, making the surface a ratio of two weighted sums rather than a single polynomial expression. 
Researchers adopt this method for leaf modeling for various plants such as cucumber, tea,  \cite{wang2011study, wu2021parametric}. 
\cite{piegl2012nurbs} define the NURBS surface as:
\begin{equation}  \mathbf{S}(u,v) = \frac{\sum_{i=0}^n \sum_{j=0}^m N_{i,p}(u) N_{j,q}(v) ~ w_{ij} \mathbf{P}_{ij}}{\sum_{i=0}^n \sum_{j=0}^m N_{i,p}(u) N_{j,q}(v)~ w_{ij}}\end{equation} 
 
where $\mathbf{P}_{ij}$ are the control point vertices, $w_{ij} > 0$ are the rational weights, and $
N_{i,p}(u) N_{j,q}(v)$ are the same Cox-de Boor basis functions as in the B-spline formulation. 
The denominator normalizes the weighted sum so that the basis functions summed to one at every point on the surface.
When all the weights are equal, the expression reduces exactly to a polynomial B-spline. 
Increasing the weight of a control point pulls the surface closer to it, wile decreasing the weight pushes the surface away. 
This gives the method a degree of local shape control that polynomial B-splines does not have. 
Rather than matching every data point exactly, the method finds the control net that minimizes the total squared distance between surface and all data points. This averaging effect means isolated noisy measurements have less influence on the final surface, making the approximation more robust to noise than exact interpolation. The rational weights then provide an additional layer of shape refinement, allowing geometrically complex regions such as midribs and leaf margins to conform more closely to the data without increasing the number of control points.

\subsection{Discrete $D^2$-Spline (D2S) }
 
\cite{kempthorne2015surface} introduce the discrete smoothing $D^2$-spline to reconstruct leaf surfaces from 3D scanned point cloud data. 
Unlike B-spline and NURBS methods, which require the point cloud to be resampled onto a structured parametric grid before fitting, the $D^2$-spline works directly with the scattered/raw measurements. This makes it more appropriate  to the non-uniform sampling densities that appear near leaf margins and veins, where reliable points are sparse. 
The method finds smooth height function $w = f(u,v)$ by solving an optimization problem that balances two competing objectives \cite{kempthorne2015surface}:

\begin{equation} \begin{split} f = \arg \min_{f \in C^1(\Omega)} \sum_{i=1}^M (w_i - f(u_i, v_i))^2 + \alpha \int_\Omega \left(\frac{\partial^2f}{\partial u^2}\right)^2  \\  +\left(\frac{\partial^2f}{\partial u \partial v}\right)^2 +\left(\frac{\partial^2f}{\partial v^2}\right)^2 dudv \end{split} \end{equation}

The first term measures how closely the surface fits the $M$ scattered data points. The second term measures the total bending energy of the surface by integrating the three independent second-order curvature components over the entire domain. 
A surface that bends sharply to follow noise receives a higher penalty from this term, so the optimization  naturally avoids overfitting.
The scalar $\alpha > 0$ controls the balance between these two objectives. A small $\alpha$ prioritizes fitting the data closely, while a large $\alpha $ prioritizes a smooth surface. 
Rather than requiring the user to set $\alpha$ manually, the method selects it automatically by minimizing the generalized cross-validation (GCV) criterion:
\begin{equation}  GCV(\alpha) = \frac{\frac{1}{M} \| (\mathbf{I} - \mathbf{H}_\alpha \mathbf{w}\|^2}{\left(\frac{1}{M}  tr(\mathbf{I} - \mathbf{H}_\alpha \mathbf{w}) \right)^2}\end{equation} 
Here, $\mathbf{H}_\alpha$ maps observations to fitted values, with $\alpha$ chosen via GCV to minimize prediction error. 
Expressing $f$ in basis functions yields a linear system combining data fitting and bending energy. $D^2$-splines achieve smoother surfaces and faster convergence than thin-plate and radial basis methods due to curvature regularization and automatic $\alpha $ selection \cite{kempthorne2015surface}.

\subsection{Delaunay triangulation } 
Delaunay triangulation creates a triangular mesh from a set of points by connecting points in a specific way \cite{de2008computational}.
A triangulation is called Delaunay triangulation if it satisfies the empty circle property: for any triangle in the mesh, the circle that passes through its three vertices does not contain any other points inside it. 
This property ensures that the resulting triangulation avoids  
silver triangles (i.e., very small triangles- close to $0^\circ$) and produces well-shaped triangles.

For leaf surface reconstruction, existing approaches project the leaf points into a plane and apply Delaunay triangulation \cite{field1988laplacian}.
 \citet{uol2024laststraw} applied 2.5D Delaunay triangulation for leaf surface reconstruction. 
They first projected the leaf point cloud into a two-dimensional plane ignoring z-axis. Next, they applied Delaunay triangulation on the  2D coordinates and then back-projected the triangulated mesh into 3D coordinates. 

\begin{figure*}
  \centering
   \includegraphics[width=0.99\linewidth]{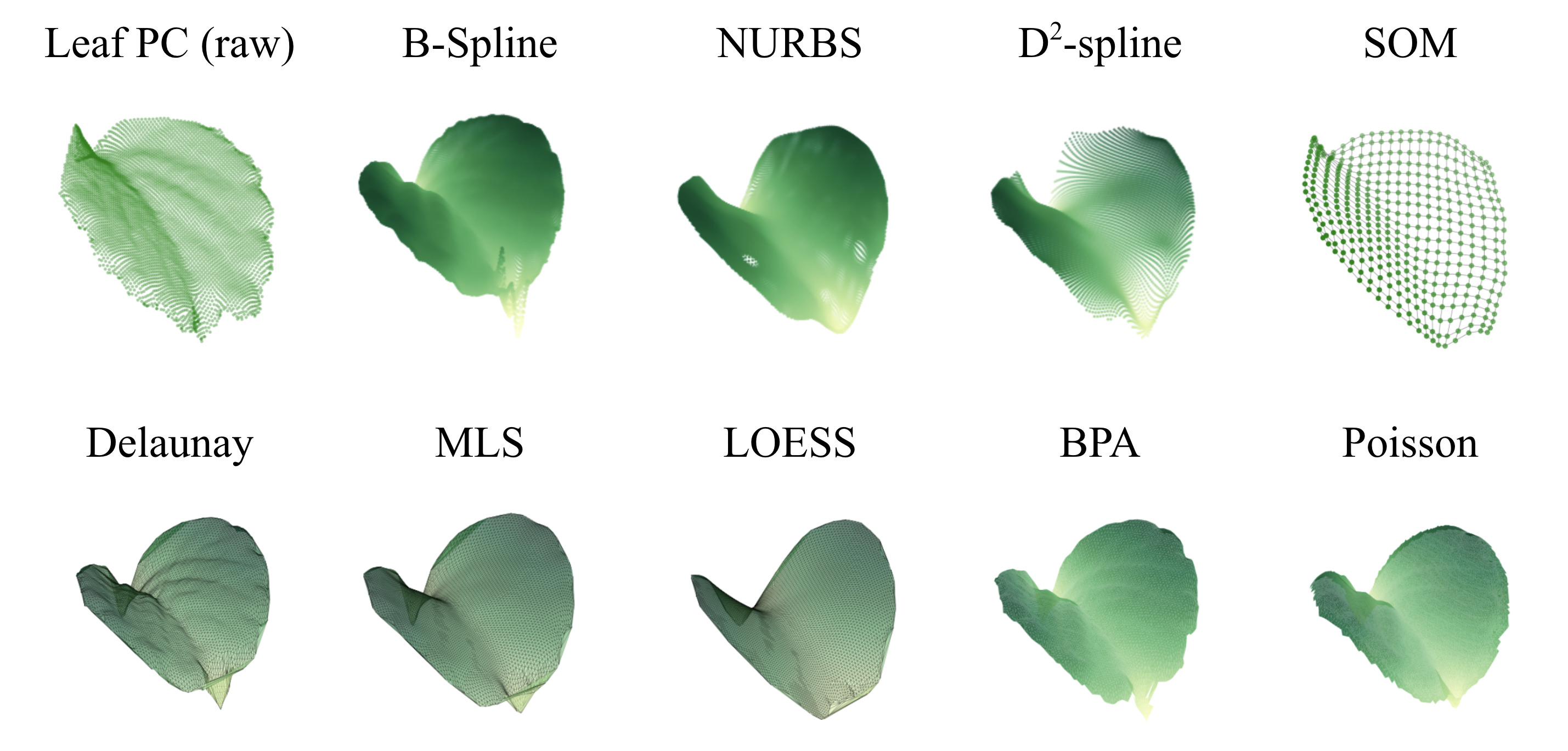} 
    \caption{Leaf surface Reconstruction
    results 
    comparison across methods that we described in Sec.~\ref{sec:all_lsr_methods}
    }\label{fig1}
    \label{fig:leaflet_lsr_results}
\end{figure*}

\subsection{Moving Least Squares}
This is a local polynomial fitting method that approximates surfaces at any query position by minimizing weighted least squares error within a local neighborhood \cite{fleishman2005robust}.
This method generates a local reference plane at each query point, projects nearby points onto this plane. 
Then, a polynomial function is fitted to minimize the weighted sum of squared distances.
For a query point $q$, MLS first finds a local reference plane $H$ with normal $n$ at position $r$. The method minimizes:

\begin{equation}  \sum_{i=1}^N  (g(x-I, y_i) - f_i)^2 \theta(\|p_i - q\|)\end{equation} 
where $f_i$ is the height of point $p_i$ over the plane $H$, defined as $f_i = n \cdot (p_i -q)$, and $g(x_i, y_i)$ is the polynomial approximation on the plane. 
Here, a distance-based gaussian kernel function assigns larger weights to closer points:
\begin{equation}  \theta(d) = e^{-\frac{d^2}{h^2}}\end{equation} 
where $d = \|p_i-q\|$ and $h$ is the bandwidth parameter. 
This method handles varying point densities for leaf surface reconstruction and provides inherent smoothing. 
 
 \subsection{Locally estimated scatterplot smoothing (LOESS) }
\citet{zhu20183d} used LOESS \cite{cleveland1988locally} instead of MLS for leaf surface fitting of maize and rice plants. Although LOESS is similar to MLS, this method works better when there are missing or sparse leaf points. The use of a tri-cube weight function makes it robust to outliers.

LOESS statistically reconstructs the surface without constructing a reference plane, making it more flexible for discontinuous data. For a query point, LOESS identifies the neighbors and computes weights using a tri-cube kernel:
\begin{equation}  W(u) = (1 - u^3)^3 \text{ for } 0 \leq u < 1, \text{ and } 0 \text{ otherwise} \end{equation} 

where $u=\rho(x, x_i)/d(x)$, with $\rho(x, x_i)$ is the distance from the query point to neighbor $x_i$, and $d(x)$ is the distance to the nearest neighbor. 
The weight for each neighbor is then $w_i(x)= W(\rho(x, x_i)/d(x))$.
LOESS can reconstruct  a continuous surface even with presence of discontinuity of leaf points, as it adapts locally. 

\subsection{Poisson Surface Reconstruction} This reconstruction method first converts oriented point clouds into watertight mesh surfaces by solving a partial differential equation \cite{kazhdan2006poisson}. 
The method takes the 3D points with outward-facing normal vectors $\{\mathbf{p}_i, \mathbf{n}_i\}$and reframes reconstruction as a mathematical problem. 
It computes an indicator function $\chi: \mathbb{R}^3 \to [0,1] $ that distinguishes the object's interior and exterior. 
The gradient of this indicator function should match a vector field $\mathbf{V}$ constructed from the input normals:
\begin{equation}  \nabla \chi = \mathbf{V}\end{equation} 
Taking the divergence of both sides yields Poisson's equation:
\begin{equation}  \Delta \chi = \nabla \cdot \mathbf{V}\end{equation} 

where $\Delta = \nabla^2 $ is the Laplacian operator. 
The algorithm solves this equation to find the smooth scalar field $\chi$ that best satisfies the normal constraints (i.e., least-squares). 
The final surface is extracted as an iso-contour where the indicator function reaches a constant value, representing the boundary between interior and exterior.
  \begin{equation}    \partial M = \{\mathbf{x} \in \mathbb{R}^3: \chi(\mathbf{x}) = c \}\end{equation}  
  
Typically, with $c=0.5$ representing the boundary between interior  ($\chi = 1$) and exterior ($\chi=0$). 
This method handles noise and produces watertight mesh without gaps even from imperfect or incomplete point cloud data.
However, \citet{zhu20183d} reported in their experimental results that the Poisson surface reconstruction method produces bumpy surfaces in the presence of noisy point cloud data for the leaves of maize plant.

\subsection{Ball Pivoting Algorithm (BPA)} This surface reconstruction method creates triangular meshes from point cloud by simulating a ball of radius $r$ rolling over points \cite{bernardini2002ball}.
The algorithm works by pivoting a virtual ball of radius $r$ across the point cloud, where the radius determines the level of detail in the final mesh. 
The ball starts from a seed triangle and expands the mesh front by pivoting around boundary edges.

For an existing edge $\sigma = (v_i, v_j)$ on the mesh front, the ball pivots to find a point $v_k$ such that the ball of radius $r$ touches all three points $v_i$, $v_j$, and $v_k$ without enclosing any other points. The center $\mathbf{c}$ of the ball satisfies:
\begin{equation}  \|\mathbf{c}- v_i\| = \| \mathbf{c}-v_j\| = \|\mathbf{c}- v_k\|=r \end{equation} 
and for all other points $v_l \notin \{v_i, v_j, v_k\}$:
\begin{equation}  \|\mathbf{c}- v_l\| \geq r\end{equation} 

When such a point $v_k$ is found, the triangle $(v_i, v_j, v_k)$ is added to the mesh, and new edges are added to the front. 
The ball continues to pivot around frontier edges until no more triangles can be formed.

However, \citet{uol2024laststraw} showed in their leaf reconstruction analysis that BPA fails when point sampling density is uneven, leaving part of leaf regions disconnected.

\subsection{Self-organizing Map (SOM)}
\citet{zermas2017estimating} use Self-Organizing Map (SOM) to reconstruct the leaf surfaces from 3D point cloud. SOM was introduced by Kohonen \cite{kohonen1997exploration}, this algorithm uses competitive learning where neurons in a 2D lattice adapt their positions in 3D space to match the point cloud geometry, with winning neurons and their neighbors adjusting to capture the leaf surface structure. 
Given a 2D lattice of neurons with weight vectors $\mathbf{m}_i \in \mathbb{R}^3$ for $i=1,.. M$, and input points $\mathbf{x}(t)$ from the point cloud, the algorithm iteratively updates neuron positions. For each input point, the best-matching cell (i.e., winner neuron) $c$ is identified based on Euclidean distance:
\begin{equation}  c = argmin_i \|\mathbf{x}(t)- \mathbf{m}_i(t) \|\end{equation} 
The weight vectors are then updated according to:
\begin{equation}  \mathbf{m}_i(t+1) = \mathbf{m}_i(t) + h_{ci}(t) [\mathbf{x}(t) - m_i(t)]\end{equation} 
where $h_{ci}(t)$ is the neighborhood kernel function. For neurons within the neighborhood $N_c(t)$ of the winner, the kernel is defined as:
\begin{equation}  h_{ci}(t) = \alpha(t), ~\textbf{if }~ i \in N_c(t)\end{equation}  and $h_{ci}(t) =0$ otherwise. More specifically, we used a Gaussian neighborhood kernel. 
%
Interestingly, SOM naturally handles varying point densities and accurately captures leaf geometry without requiring the uniform sampling density like BPA algorithm. 
Although SOM can reconstruct leaf surfaces effectively, their computational complexity hinders deployment in real-time robotic pipeline. 
We therefore  propose a learning based leaf surface reconstruction method that achieves comparable accuracy with significantly improved inference speed that can support real-time robotics applications such as the phenotyping task.


\section{Performance of Leaf Surface Reconstruction Methods}
\label{sec:3_performance}

\begin{table*}[!htbp]
\centering
\caption{Surface reconstruction results for ten Leaves across nine plant species from Crops3d ( $\mathcal{D}_{C3D}$) and Pheno4D (\textcolor{black}{ $\mathcal{D}_{P4D}$}) Datasets }
\label{tbl:c3d_p4d}

\renewcommand\arraystretch{1.0}

\begin{adjustbox}{width=\textwidth}
\begin{tabular}{c c c c c c c c c c} %
\toprule
\textbf{Leaf {\footnotesize Point Cloud} }
& \textbf{B-spline }
& \textbf{NURBS} 
& \textbf{$D^2$-spline} 
& \textbf{Delaunay} 
& \textbf{MLS} 
& \textbf{LOESS} 
& \textbf{BPA} 
& \textbf{Poisson}  
& \textbf{SOM}  \\

\midrule

\includegraphics[width=0.08\textwidth]{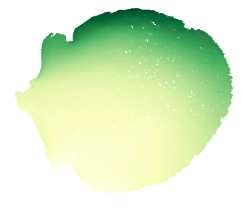}
& \includegraphics[width=0.08\textwidth ]{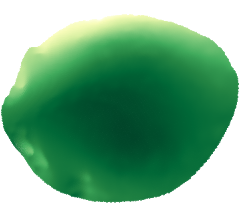} 
& \includegraphics[width=0.08\textwidth ]{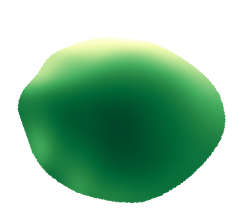} 
& \includegraphics[width=0.08\textwidth ]{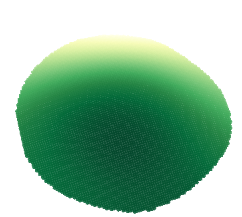} 
& \includegraphics[width=0.08\textwidth ]{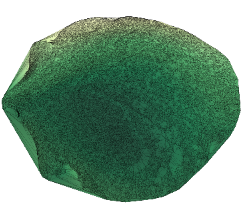} 
& \includegraphics[width=0.08\textwidth ]{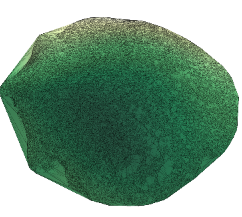} 
& \includegraphics[width=0.08\textwidth ]{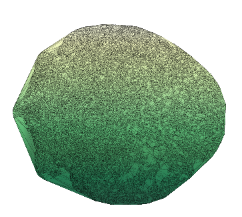} 
& \includegraphics[width=0.08\textwidth ]{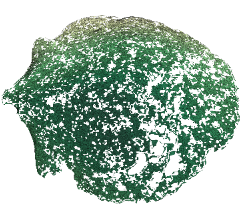} 
& \includegraphics[width=0.08\textwidth ]{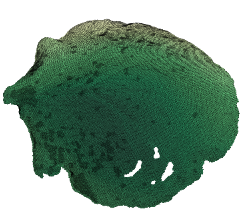} 
& \includegraphics[width=0.08\textwidth ]{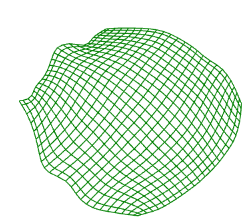} 
\\  %
 
 {\footnotesize \makecell{  Cabbage {\tiny $\mathcal{D}_{C3D}$ (47K points)}   \\  CPU Time \\ RAM Usage}}
& {\footnotesize   \makecell{  2519   {\tiny $\mathrm{Unit}^2$} \\ 44.1{\tiny S} \\ 25.9{\tiny MB}} } 
& {\footnotesize   \makecell{  2429   {\tiny $\mathrm{Unit}^2$} \\ 24.2{\tiny S} \\ 21.9{\tiny MB}} }
& {\footnotesize   \makecell{  2269   {\tiny $\mathrm{Unit}^2$} \\ 6.06{\tiny S} \\ 53.6{\tiny MB}} } 
& {\footnotesize   \makecell{  2518   {\tiny $\mathrm{Unit}^2$} \\ 32.9{\tiny S} \\ 275.6{\tiny MB}} }
& {\footnotesize   \makecell{  2396   {\tiny $\mathrm{Unit}^2$} \\ 617.6{\tiny S} \\ 896.9{\tiny MB}} }

& {\footnotesize   \makecell{  2382   {\tiny $\mathrm{Unit}^2$} \\ 34.1{\tiny S} \\ 277.6{\tiny MB}}} 
& {\footnotesize   \makecell{  1628   {\tiny $\mathrm{Unit}^2$} \\ 27.5{\tiny S} \\ 186.6{\tiny MB}} } 
 & {\footnotesize   \makecell{  2518   {\tiny $\mathrm{Unit}^2$} \\ 61.9{\tiny S} \\ 444.9{\tiny MB}} }  
& {\footnotesize   \makecell{    1944   {\tiny $\mathrm{Unit}^2$} \\  779.9{\tiny S} \\  2.8{\tiny MB}}} 

\\
\midrule

\includegraphics[width=0.08\textwidth]{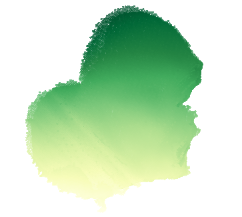}
& \includegraphics[width=0.08\textwidth ]{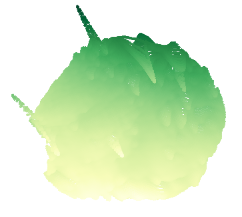} 
& \includegraphics[width=0.08\textwidth ]{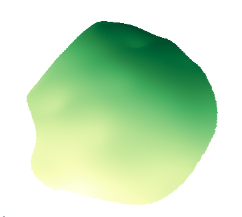} 
& \includegraphics[width=0.08\textwidth ]{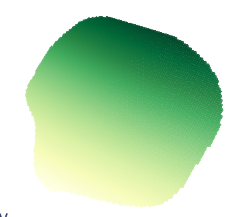} 
& \includegraphics[width=0.08\textwidth ]{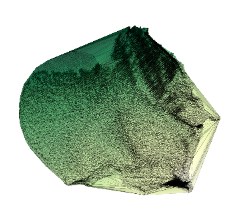} 
& \includegraphics[width=0.08\textwidth ]{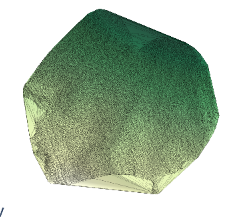} 
& \includegraphics[width=0.08\textwidth ]{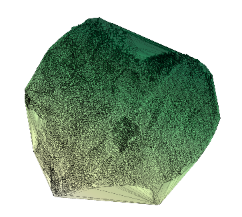} 
& \includegraphics[width=0.08\textwidth ]{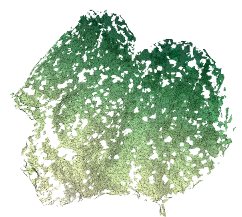} 
& \includegraphics[width=0.08\textwidth ]{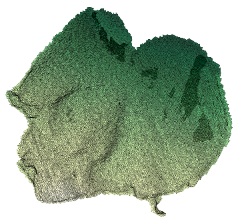} 
& \includegraphics[width=0.08\textwidth ]{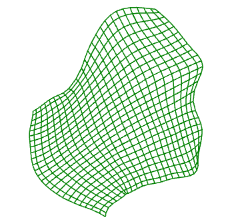} 
\\  %
 
 {\footnotesize \makecell{  Cotton {\tiny  $\mathcal{D}_{C3D}$ (46K points)}   \\  CPU Time \\ RAM Usage }}
& {\footnotesize   \makecell{    1060   {\tiny $\mathrm{Unit}^2$} \\  44.1{\tiny S} \\  26.1{\tiny MB}} } 
& {\footnotesize   \makecell{    630   {\tiny $\mathrm{Unit}^2$} \\  24.2{\tiny S} \\ 22.1{\tiny MB}} }
& {\footnotesize   \makecell{    560   {\tiny $\mathrm{Unit}^2$} \\   6.4{\tiny S} \\ 54.1{\tiny MB}} } 
& {\footnotesize   \makecell{    2090   {\tiny $\mathrm{Unit}^2$} \\  31.4{\tiny S} \\ 264.7{\tiny MB}} }
& {\footnotesize   \makecell{    620   {\tiny $\mathrm{Unit}^2$} \\  627.2{\tiny S} \\ 803.9{\tiny MB}} }
& {\footnotesize   \makecell{    1860   {\tiny $\mathrm{Unit}^2$} \\  32.3{\tiny S} \\ 266.7{\tiny MB}} }
& {\footnotesize   \makecell{   530   {\tiny $\mathrm{Unit}^2$} \\  17.2{\tiny S} \\ 56.0{\tiny MB}} } 
 & {\footnotesize   \makecell{  730    {\tiny $\mathrm{Unit}^2$} \\  45.58{\tiny S} \\ 337.5{\tiny MB}} }  
& {\footnotesize   \makecell{    447   {\tiny $\mathrm{Unit}^2$} \\  711.8 {\tiny S} \\ 2.0 {\tiny MB}}} 

\\
\midrule

\includegraphics[width=0.08\textwidth]{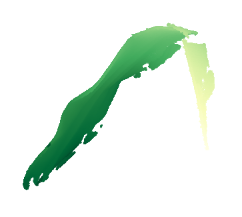}
& \includegraphics[width=0.08\textwidth ]{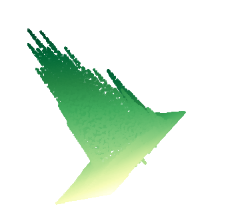} 
& \includegraphics[width=0.08\textwidth ]{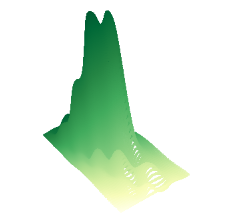} 
& \includegraphics[width=0.08\textwidth ]{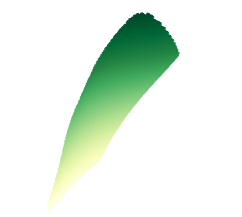} 
& \includegraphics[width=0.08\textwidth ]{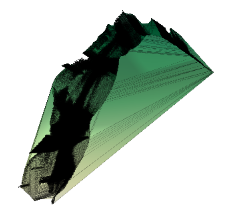} 
& \includegraphics[width=0.08\textwidth ]{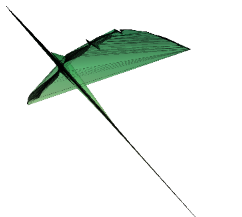} 
& \includegraphics[width=0.08\textwidth ]{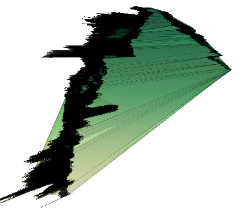} 
& \includegraphics[width=0.08\textwidth ]{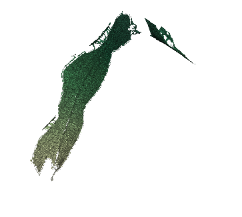} 
& \includegraphics[width=0.08\textwidth ]{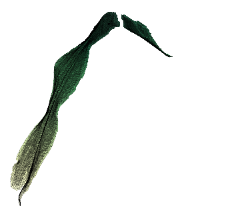} 
& \includegraphics[width=0.08\textwidth ]{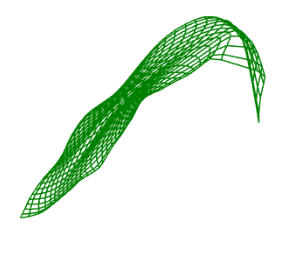} 

\\  %

 {\footnotesize \makecell{  Maize {\tiny  $\mathcal{D}_{C3D}$ (43K points)}   \\  CPU Time \\ RAM Usage  }}
& {\footnotesize   \makecell{    256961.4 {\tiny $\mathrm{Unit}^2$} \\  48.4{\tiny S} \\ 28.2{\tiny MB}} } 
& {\footnotesize   \makecell{    150150.1 {\tiny $\mathrm{Unit}^2$} \\  26.8{\tiny S} \\ 24.5{\tiny MB}} }
& {\footnotesize   \makecell{    8265.9 {\tiny $\mathrm{Unit}^2$} \\   5.6{\tiny S} \\ 43.9{\tiny MB}} } 
& {\footnotesize   \makecell{    25028.6 {\tiny $\mathrm{Unit}^2$} \\  30.4{\tiny S} \\ 253.6{\tiny MB}} }
& {\footnotesize   \makecell{    126076.1 {\tiny $\mathrm{Unit}^2$} \\  841.5{\tiny S} \\ 1918.4{\tiny MB}} }
& {\footnotesize   \makecell{    14899.5 {\tiny $\mathrm{Unit}^2$} \\  30.0{\tiny S} \\ 255.5{\tiny MB}} }
& {\footnotesize   \makecell{    3233.7 {\tiny $\mathrm{Unit}^2$} \\  22.5{\tiny S} \\ 160.2{\tiny MB}} } 
 & {\footnotesize   \makecell{    5357.2 {\tiny $\mathrm{Unit}^2$} \\  41.8{\tiny S} \\ 313.329{\tiny MB}} }  
& {\footnotesize   \makecell{   3148   {\tiny $\mathrm{Unit}^2$} \\  662.9{\tiny S} \\ 2.8{\tiny MB}}} 

\\
\midrule

\includegraphics[width=0.08\textwidth]{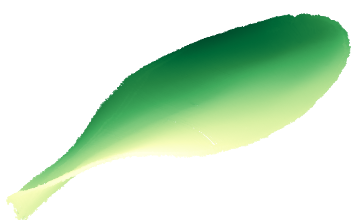}
& \includegraphics[width=0.08\textwidth ]{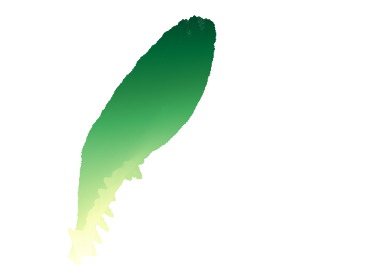} 
& \includegraphics[width=0.08\textwidth ]{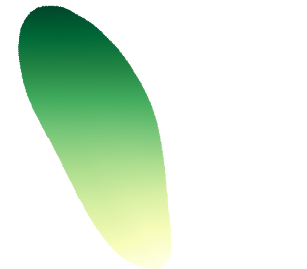} 
& \includegraphics[width=0.08\textwidth ]{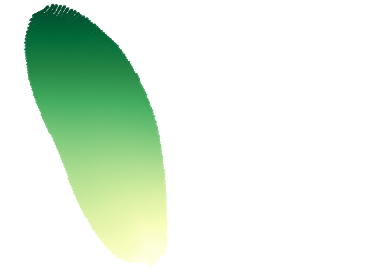} 
& \includegraphics[width=0.08\textwidth ]{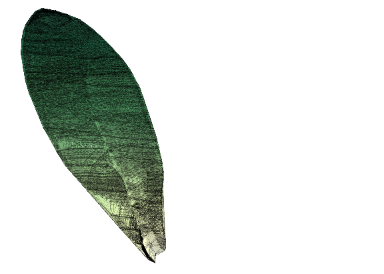} 
& \includegraphics[width=0.08\textwidth ]{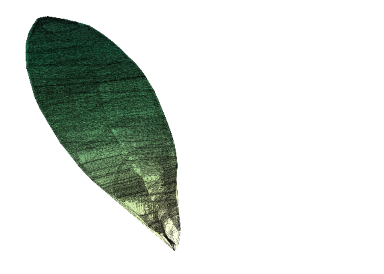} 
& \includegraphics[width=0.08\textwidth ]{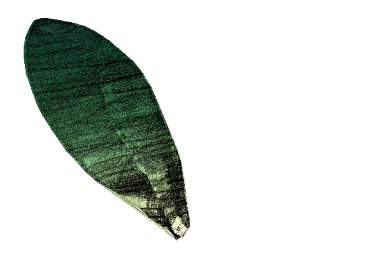} 
& \includegraphics[width=0.08\textwidth ]{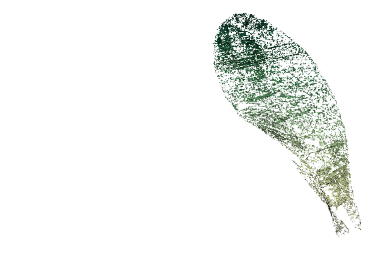} 
& \includegraphics[width=0.08\textwidth ]{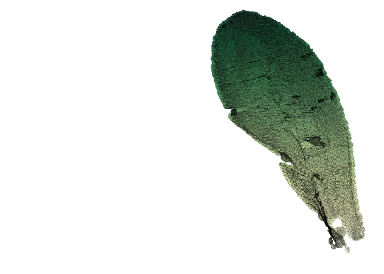} 
& \includegraphics[width=0.08\textwidth ]{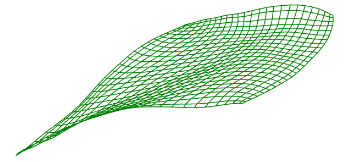} 
\\  %

 {\footnotesize \makecell{  Maize {\tiny \textcolor{black}{\textbf{ $\mathcal{D}_{P4D}$}} (156K points)}     \\  CPU Time \\ RAM Usage }}
& {\footnotesize   \makecell{   612.2   {\tiny $\mathrm{Unit}^2$} \\  47.7{\tiny S} \\ 60.0{\tiny MB}} } 
& {\footnotesize   \makecell{   483.3   {\tiny $\mathrm{Unit}^2$} \\  26.2{\tiny S} \\ 45.0{\tiny MB}} }
& {\footnotesize   \makecell{   510.0    {\tiny $\mathrm{Unit}^2$} \\    5.3{\tiny S} \\ 46.1{\tiny MB}} } 
& {\footnotesize   \makecell{   1134.6   {\tiny $\mathrm{Unit}^2$} \\  107.9{\tiny S} \\ 903.2{\tiny MB}} }
& {\footnotesize   \makecell{    473.2   {\tiny $\mathrm{Unit}^2$} \\  7052.3{\tiny S} \\ 31283.2.{\tiny MB}} }
& {\footnotesize   \makecell{    1422.9   {\tiny $\mathrm{Unit}^2$} \\ 110.9{\tiny S} \\ 910.1{\tiny MB}} }
& {\footnotesize   \makecell{    230.5   {\tiny $\mathrm{Unit}^2$} \\  48.7{\tiny S} \\ 153.3{\tiny MB}} } 
 & {\footnotesize   \makecell{   570.8   {\tiny $\mathrm{Unit}^2$} \\  78.2{\tiny S} \\ 546.8{\tiny MB}} }  
& {\footnotesize   \makecell{    364.9   {\tiny $\mathrm{Unit}^2$} \\  2465.3 {\tiny S} \\ 2.1{\tiny MB}}} 

\\
\midrule

\includegraphics[width=0.08\textwidth]{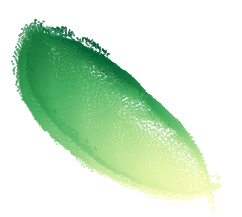}
& \includegraphics[width=0.08\textwidth ]{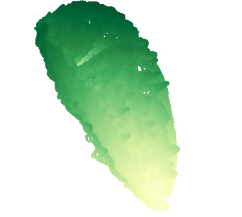} 
& \includegraphics[width=0.08\textwidth ]{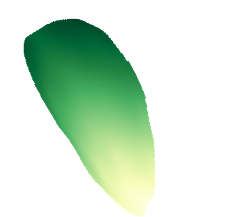} 
& \includegraphics[width=0.08\textwidth ]{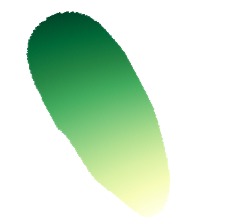} 
& \includegraphics[width=0.08\textwidth ]{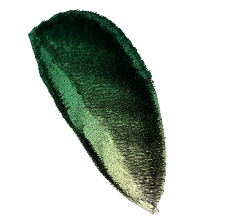} 
& \includegraphics[width=0.08\textwidth ]{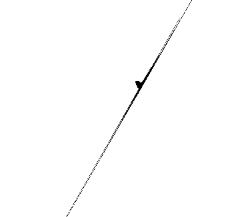} 
& \includegraphics[width=0.08\textwidth ]{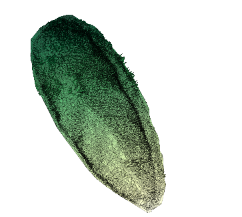} 
& \includegraphics[width=0.08\textwidth ]{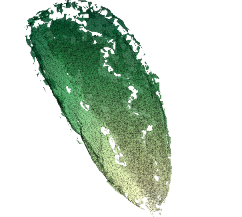} 
& \includegraphics[width=0.08\textwidth ]{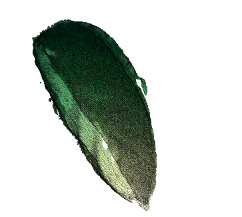} 
& \includegraphics[width=0.08\textwidth ]{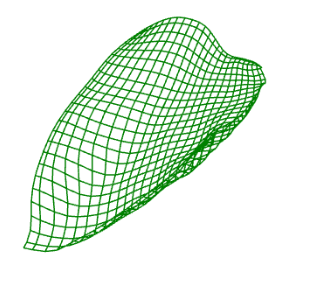} 
\\  %

 { \footnotesize \makecell{  Potato {\tiny  $\mathcal{D}_{C3D}$ (23K points)}   \\  CPU Time \\ RAM Usage }}
& {\footnotesize   \makecell{    725.3  {\tiny $\mathrm{Unit}^2$} \\  46.5{\tiny S} \\ 26.3{\tiny MB}} } 
& {\footnotesize   \makecell{    267.7  {\tiny $\mathrm{Unit}^2$} \\  25.2{\tiny S} \\ 22.6{\tiny MB}} }
& {\footnotesize   \makecell{    202.9 {\tiny $\mathrm{Unit}^2$} \\   5.3{\tiny S} \\ 43.1{\tiny MB}} } 
& {\footnotesize   \makecell{    2083.2 {\tiny $\mathrm{Unit}^2$} \\  15.88{\tiny S} \\ 133.5{\tiny MB}} }
& {\footnotesize   \makecell{    7881.0 {\tiny $\mathrm{Unit}^2$} \\  331.2{\tiny S} \\ 310.4{\tiny MB}} }
& {\footnotesize   \makecell{    1188.4 {\tiny $\mathrm{Unit}^2$} \\  16.0{\tiny S} \\ 134.6{\tiny MB}} }
& {\footnotesize   \makecell{    361.4 {\tiny $\mathrm{Unit}^2$} \\  9.1{\tiny S} \\ 40.2{\tiny MB}} } 
 & {\footnotesize   \makecell{    575.0 {\tiny $\mathrm{Unit}^2$} \\  43.9{\tiny S} \\ 332.2{\tiny MB}} }  
& {\footnotesize   \makecell{    240.2   {\tiny $\mathrm{Unit}^2$} \\  353.5{\tiny S} \\ 2.0{\tiny MB}}} 

\\
\midrule

\includegraphics[width=0.08\textwidth]{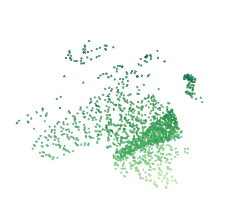}
& \includegraphics[width=0.08\textwidth ]{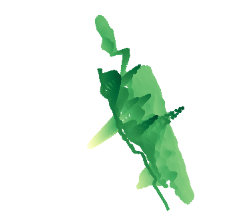} 
& \includegraphics[width=0.08\textwidth ]{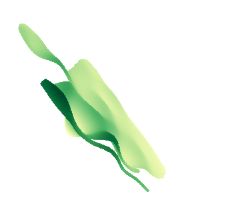} 
& \includegraphics[width=0.08\textwidth ]{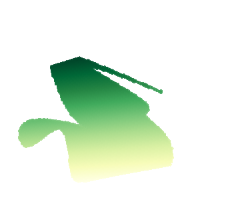} 
& \includegraphics[width=0.08\textwidth ]{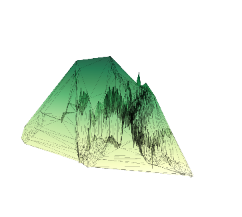} 
& \includegraphics[width=0.08\textwidth ]{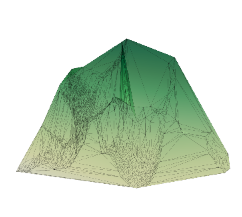} 
& \includegraphics[width=0.08\textwidth ]{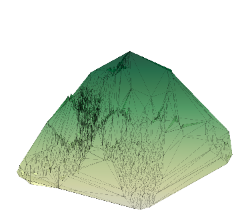} 
& \includegraphics[width=0.08\textwidth ]{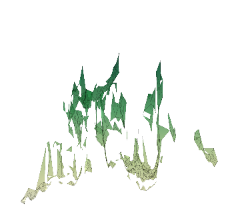} 
& \includegraphics[width=0.08\textwidth ]{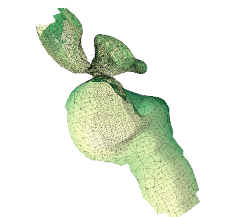} 
& \includegraphics[width=0.08\textwidth ]{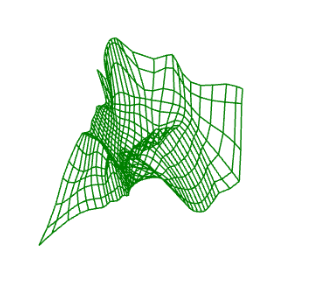} 
\\  %

 {\footnotesize \makecell{  Rapeseed {\tiny  $\mathcal{D}_{C3D}$ (1K points)}   \\  CPU Time \\ RAM Usage }}
& {\footnotesize   \makecell{    222.9   {\tiny $\mathrm{Unit}^2$} \\  42.6{\tiny S} \\ 22.1{\tiny MB}} } 
& {\footnotesize   \makecell{    116.3   {\tiny $\mathrm{Unit}^2$} \\  21.4{\tiny S} \\ 18.3{\tiny MB}} }
& {\footnotesize   \makecell{    74.6   {\tiny $\mathrm{Unit}^2$} \\   2.3{\tiny S} \\ 38.9{\tiny MB}} } 
& {\footnotesize   \makecell{    155.4   {\tiny $\mathrm{Unit}^2$} \\  1.3{\tiny S} \\ 10.9{\tiny MB}} }
& {\footnotesize   \makecell{    100.2   {\tiny $\mathrm{Unit}^2$} \\  23.1{\tiny S} \\ 11.5{\tiny MB}} }
& {\footnotesize   \makecell{    107.9   {\tiny $\mathrm{Unit}^2$} \\  1.5{\tiny S} \\ 11.1{\tiny MB}} }
& {\footnotesize   \makecell{    34.9   {\tiny $\mathrm{Unit}^2$} \\  1.1{\tiny S} \\ 3.4{\tiny MB}} } 
 & {\footnotesize   \makecell{    267.7   {\tiny $\mathrm{Unit}^2$} \\  5.3{\tiny S} \\ 36.9{\tiny MB}} }  
& {\footnotesize   \makecell{    69.9   {\tiny $\mathrm{Unit}^2$} \\  25.5{\tiny S} \\ 1.9{\tiny MB}}} 

\\
\midrule

\includegraphics[width=0.06\textwidth]{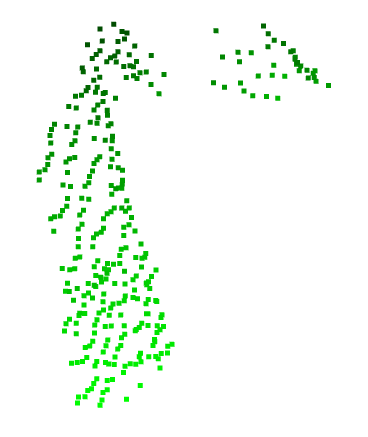}
& \includegraphics[width=0.08\textwidth ]{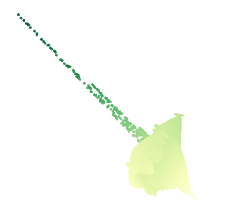} 
& \includegraphics[width=0.08\textwidth ]{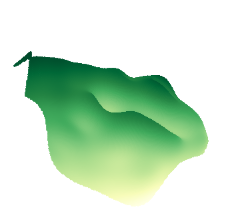} 
& \includegraphics[width=0.08\textwidth ]{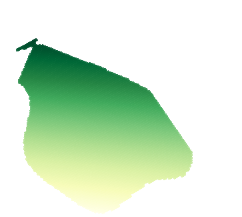} 
& \includegraphics[width=0.08\textwidth ]{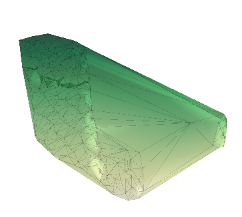} 
& \includegraphics[width=0.08\textwidth ]{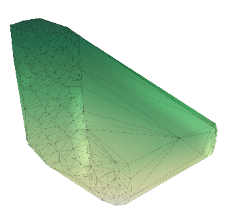} 
& \includegraphics[width=0.08\textwidth ]{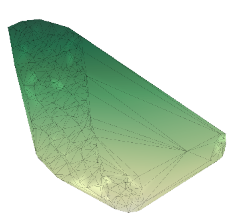} 
& \includegraphics[width=0.08\textwidth ]{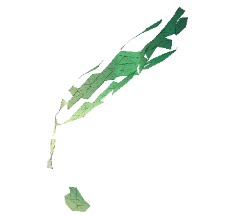} 
& \includegraphics[width=0.08\textwidth ]{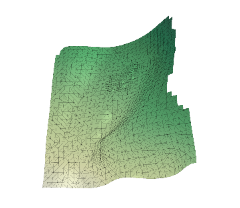} 
& \includegraphics[width=0.08\textwidth ]{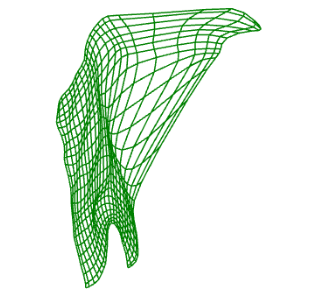} 
\\  %

 {\footnotesize \makecell{  Rice {\tiny  $\mathcal{D}_{C3D}$ (0.3K points)}   \\  CPU Time \\ RAM Usage }}
& {\footnotesize   \makecell{   219.4   {\tiny $\mathrm{Unit}^2$} \\  44.7{\tiny S} \\ 24.4{\tiny MB}} } 
& {\footnotesize   \makecell{   96.8   {\tiny $\mathrm{Unit}^2$} \\  25.2{\tiny S} \\ 20.7{\tiny MB}} }
& {\footnotesize   \makecell{   57.70   {\tiny $\mathrm{Unit}^2$} \\   2.4{\tiny S} \\ 40.3{\tiny MB}} } 
& {\footnotesize   \makecell{   71.0   {\tiny $\mathrm{Unit}^2$} \\  0.5{\tiny S} \\ 3.0{\tiny MB}} }
& {\footnotesize   \makecell{   64.3   {\tiny $\mathrm{Unit}^2$} \\  4.1{\tiny S} \\ 3.5{\tiny MB}} }
& {\footnotesize   \makecell{   58.9   {\tiny $\mathrm{Unit}^2$} \\  0.5{\tiny S} \\ 3.0{\tiny MB}} }
& {\footnotesize   \makecell{   19.3   {\tiny $\mathrm{Unit}^2$} \\  0.3{\tiny S} \\ 1.9{\tiny MB}} } 
 & {\footnotesize   \makecell{   157.1   {\tiny $\mathrm{Unit}^2$} \\  1.8{\tiny S} \\ 10.3{\tiny MB}} }  
& {\footnotesize   \makecell{   37.8   {\tiny $\mathrm{Unit}^2$} \\  5.2{\tiny S} \\ 2.2{\tiny MB}}} 

\\
\midrule

\includegraphics[width=0.08\textwidth]{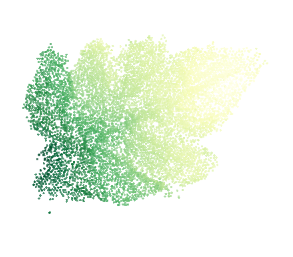}
& \includegraphics[width=0.08\textwidth ]{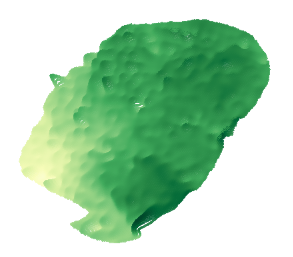} 
& \includegraphics[width=0.08\textwidth ]{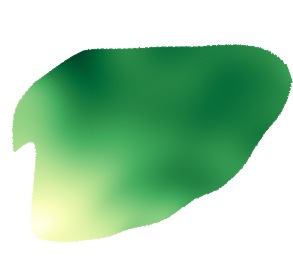} 
& \includegraphics[width=0.08\textwidth ]{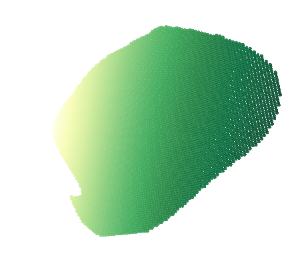} 
& \includegraphics[width=0.08\textwidth ]{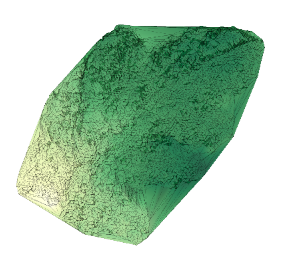} 
& \includegraphics[width=0.08\textwidth ]{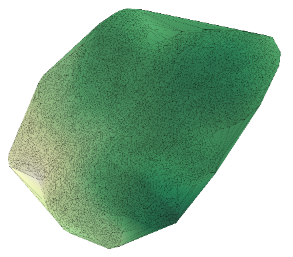} 
& \includegraphics[width=0.08\textwidth ]{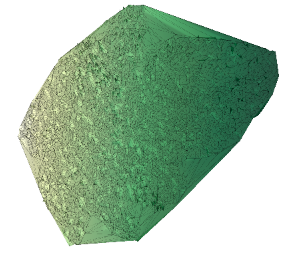} 
& \includegraphics[width=0.08\textwidth ]{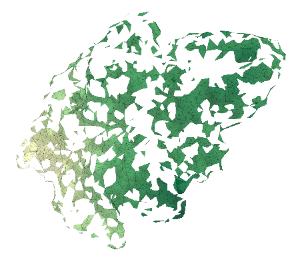} 
& \includegraphics[width=0.08\textwidth ]{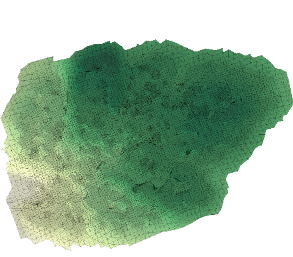} 
& \includegraphics[width=0.08\textwidth ]{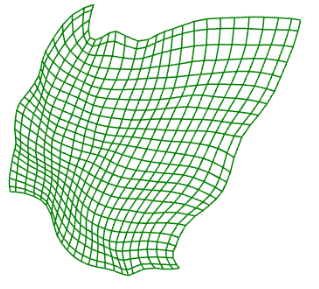} 
\\  %

 {\footnotesize \makecell{  Tomato {\tiny  $\mathcal{D}_{C3D}$ (8K points)}    \\  CPU Time \\ RAM Usage }}
& {\footnotesize   \makecell{   168.4   {\tiny $\mathrm{Unit}^2$} \\  44.4{\tiny S} \\ 24.4{\tiny MB}} } 
& {\footnotesize   \makecell{   108.5   {\tiny $\mathrm{Unit}^2$} \\  23.3{\tiny S} \\ 20.7{\tiny MB}} }
& {\footnotesize   \makecell{   101.6   {\tiny $\mathrm{Unit}^2$} \\   5.0{\tiny S} \\ 41.5{\tiny MB}} } 
& {\footnotesize   \makecell{   237.6   {\tiny $\mathrm{Unit}^2$} \\  6.1{\tiny S} \\ 50.7{\tiny MB}} }
& {\footnotesize   \makecell{   109.6   {\tiny $\mathrm{Unit}^2$} \\  89.4{\tiny S} \\ 51.9{\tiny MB}} }
& {\footnotesize   \makecell{   164.5   {\tiny $\mathrm{Unit}^2$} \\  6.2{\tiny S} \\ 51.0{\tiny MB}} }
& {\footnotesize   \makecell{   53.3   {\tiny $\mathrm{Unit}^2$} \\  2.6{\tiny S} \\ 8.4{\tiny MB}} } 
 & {\footnotesize   \makecell{   165.5   {\tiny $\mathrm{Unit}^2$} \\  13.4{\tiny S} \\ 97.3{\tiny MB}} }  
& {\footnotesize   \makecell{  79.9    {\tiny $\mathrm{Unit}^2$} \\  132.7{\tiny S} \\ 2.2{\tiny MB}}} 

\\
\midrule
\includegraphics[width=0.08\textwidth]{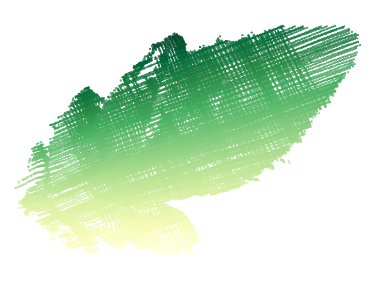}
& \includegraphics[width=0.08\textwidth ]{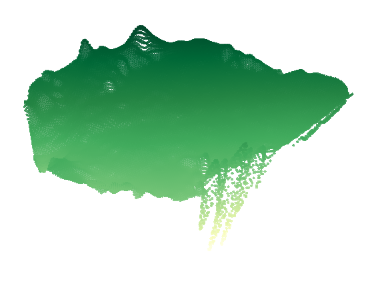} 
& \includegraphics[width=0.08\textwidth ]{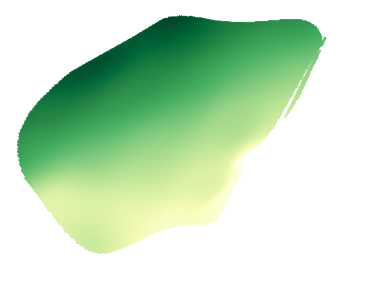} 
& \includegraphics[width=0.08\textwidth ]{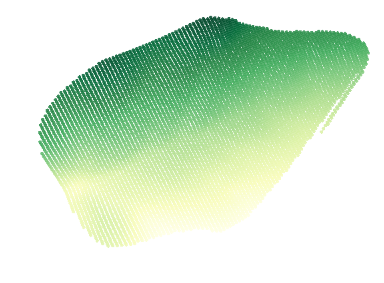} 
& \includegraphics[width=0.08\textwidth ]{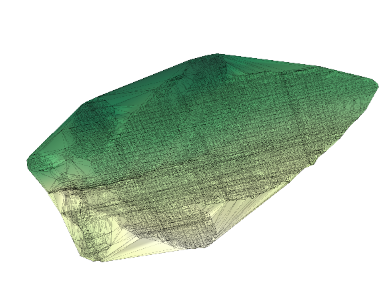} 
& \includegraphics[width=0.08\textwidth ]{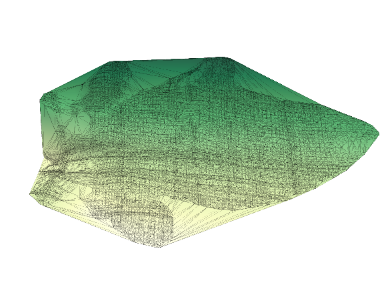} 
& \includegraphics[width=0.08\textwidth ]{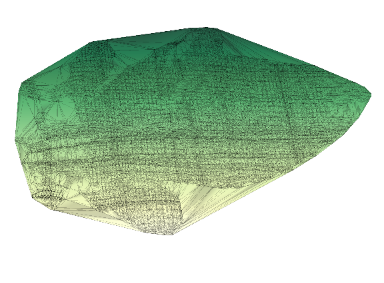} 
& \includegraphics[width=0.08\textwidth ]{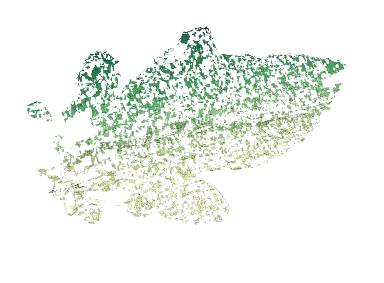} 
& \includegraphics[width=0.08\textwidth ]{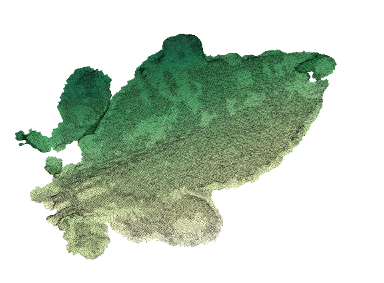} 
& \includegraphics[width=0.08\textwidth ]{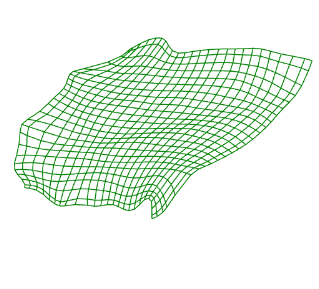} 
\\  %

 {\footnotesize \makecell{  Tomato {\tiny \textcolor{black}{\textbf{ $\mathcal{D}_{P4D}$}} (38K points)}   \\  CPU Time \\ RAM Usage }}
& {\footnotesize   \makecell{   240.4   {\tiny $\mathrm{Unit}^2$} \\  46.9{\tiny S} \\ 24.3{\tiny MB}} } 
& {\footnotesize   \makecell{   165.9   {\tiny $\mathrm{Unit}^2$} \\  23.0{\tiny S} \\ 20.6{\tiny MB}} }
& {\footnotesize   \makecell{   164.2   {\tiny $\mathrm{Unit}^2$} \\   5.15{\tiny S} \\ 42.2{\tiny MB}} } 
& {\footnotesize   \makecell{   231.7   {\tiny $\mathrm{Unit}^2$} \\  26.5{\tiny S} \\ 220.8{\tiny MB}} }
& {\footnotesize   \makecell{   164.5   {\tiny $\mathrm{Unit}^2$} \\  685.5{\tiny S} \\ 1167.5{\tiny MB}} }
& {\footnotesize   \makecell{   241.4   {\tiny $\mathrm{Unit}^2$} \\  28.3{\tiny S} \\ 222.5{\tiny MB}} }
& {\footnotesize   \makecell{   55.7   {\tiny $\mathrm{Unit}^2$} \\  9.27{\tiny S} \\ 39.2{\tiny MB}} } 
 & {\footnotesize   \makecell{    178.4   {\tiny $\mathrm{Unit}^2$} \\  66.5{\tiny S} \\ 490.4{\tiny MB}} }  
& {\footnotesize   \makecell{   116.4   {\tiny $\mathrm{Unit}^2$} \\  654.6{\tiny S} \\ 2.2{\tiny MB}}} 

\\
\midrule

\includegraphics[width=0.08\textwidth]{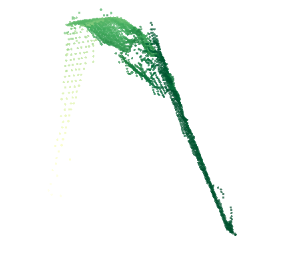}
& \includegraphics[width=0.08\textwidth ]{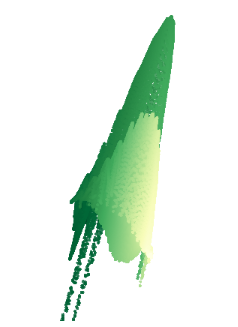} 
& \includegraphics[width=0.08\textwidth ]{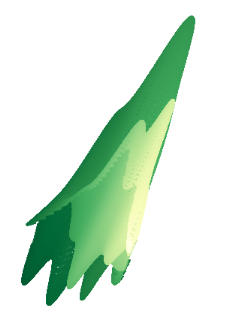} 
& \includegraphics[width=0.08\textwidth ]{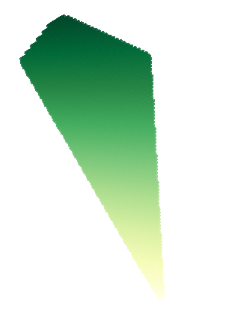} 
& \includegraphics[width=0.08\textwidth ]{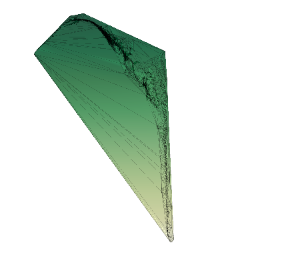} 
& \includegraphics[width=0.08\textwidth ]{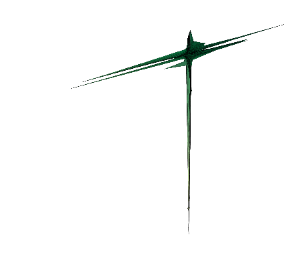} 
& \includegraphics[width=0.08\textwidth ]{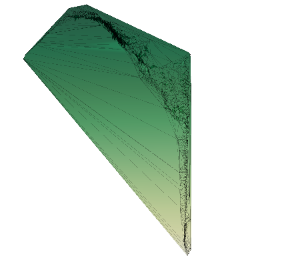} 
& \includegraphics[width=0.08\textwidth ]{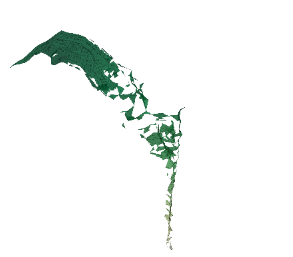} 
& \includegraphics[width=0.08\textwidth ]{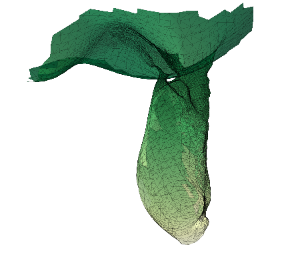} 
& \includegraphics[width=0.08\textwidth ]{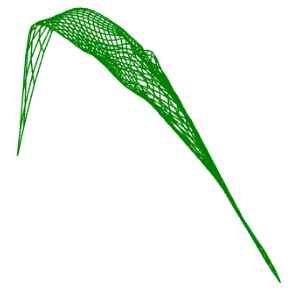} 
\\  %

 {\footnotesize \makecell{  Wheat {\tiny  $\mathcal{D}_{C3D}$ (2K points)}   \\  CPU Time \\ RAM Usage }}
& {\footnotesize   \makecell{   10209.7   {\tiny $\mathrm{Unit}^2$} \\  187.7{\tiny S} \\ 13.1{\tiny MB}} } 
& {\footnotesize   \makecell{   5731.1   {\tiny $\mathrm{Unit}^2$} \\  88.3{\tiny S} \\ 14.5{\tiny MB}} }
& {\footnotesize   \makecell{   811.9   {\tiny $\mathrm{Unit}^2$} \\   11.9{\tiny S} \\ 0.8{\tiny MB}} } 
& {\footnotesize   \makecell{   1767.9   {\tiny $\mathrm{Unit}^2$} \\  2.2{\tiny S} \\ 17.3{\tiny MB}} }
& {\footnotesize   \makecell{   9380.9   {\tiny $\mathrm{Unit}^2$} \\  28.7{\tiny S} \\ 18.3{\tiny MB}} }
& {\footnotesize   \makecell{   1263.6   {\tiny $\mathrm{Unit}^2$} \\  2.1{\tiny S} \\ 17.4{\tiny MB}} }
& {\footnotesize   \makecell{   227.0   {\tiny $\mathrm{Unit}^2$} \\  1.1.{\tiny S} \\ 6.3{\tiny MB}} } 
 & {\footnotesize   \makecell{   3637.4   {\tiny $\mathrm{Unit}^2$} \\  8.3{\tiny S} \\ 60.7{\tiny MB}} }  
& {\footnotesize   \makecell{    378.4   {\tiny $\mathrm{Unit}^2$} \\  42.9{\tiny S} \\ 2.1{\tiny MB}}} 

\\
\midrule

\bottomrule
\end{tabular}
\end{adjustbox}

\end{table*}

In this section, we evaluate the performance of leaf surface reconstruction methods on real leaf point cloud data. We first describe the dataset preparation and then describe the implementation details for each method. Next, we report the results we obtain in our experiments.

\subsection{Dataset Preparation}
\label{sec:custom_dataset}

We extract leaf point cloud samples for tomato and maize plants from Pheno4d \cite{schunck2021pheno4d} dataset, which we denote as $\mathcal{D}_{P4D}$. Pheno4D provides high-resolution point cloud for 3D plant phenotyping, which was collected in controlled greenhouse environments using structured-light RGB-D sensors.  The dataset includes detail annotation at both leaf and plant level.
We manually segment leaf point clouds of eight plants (cabbage, cotton, maize, potato, rapeseed, rice, tomato, and wheat) from 
Crops3D \cite{zhu2024crops3d}, which we represent as $\mathcal{D}_{C3D}$. 
Compared to Pheno4D, Crops3D is a real-world 3D point cloud dataset designed for agricultural phenotyping. It provides plants 3D point clouds constructed from multiple methods including LiDAR, structural light, and multi-view stereo reconstruction under natural field conditions. However, this dataset do not provide high resolution point cloud similar to Pheno4D.

The publicly available LAST-STRAW dataset contains 28 annotated strawberry point clouds~\cite{uol2024laststraw}. 
There are nearly a million data points available for these plant point clouds. These point clouds were constructed from two different plants over 11 weeks.   
However, the trifoliate structure of the clusters of strawberry leaves makes leaf reconstruction much more challenging. So, we manually segment the strawberry leaf cluster using CloudCompare, which allows us to extract 893 individual leaves (i.e., leaflets) from annotated leaf clusters, which we denote as $\mathcal{D_{STRAW}}$.\footnote{https://www.cloudcompare.org/}
Accurate leaf-surface reconstruction requires reliable leaf segmentation, and this manual process provides high-quality leaf point cloud for our analysis.

\begin{figure*}[!htbp]
    \centering
    \includegraphics[width=1\linewidth]{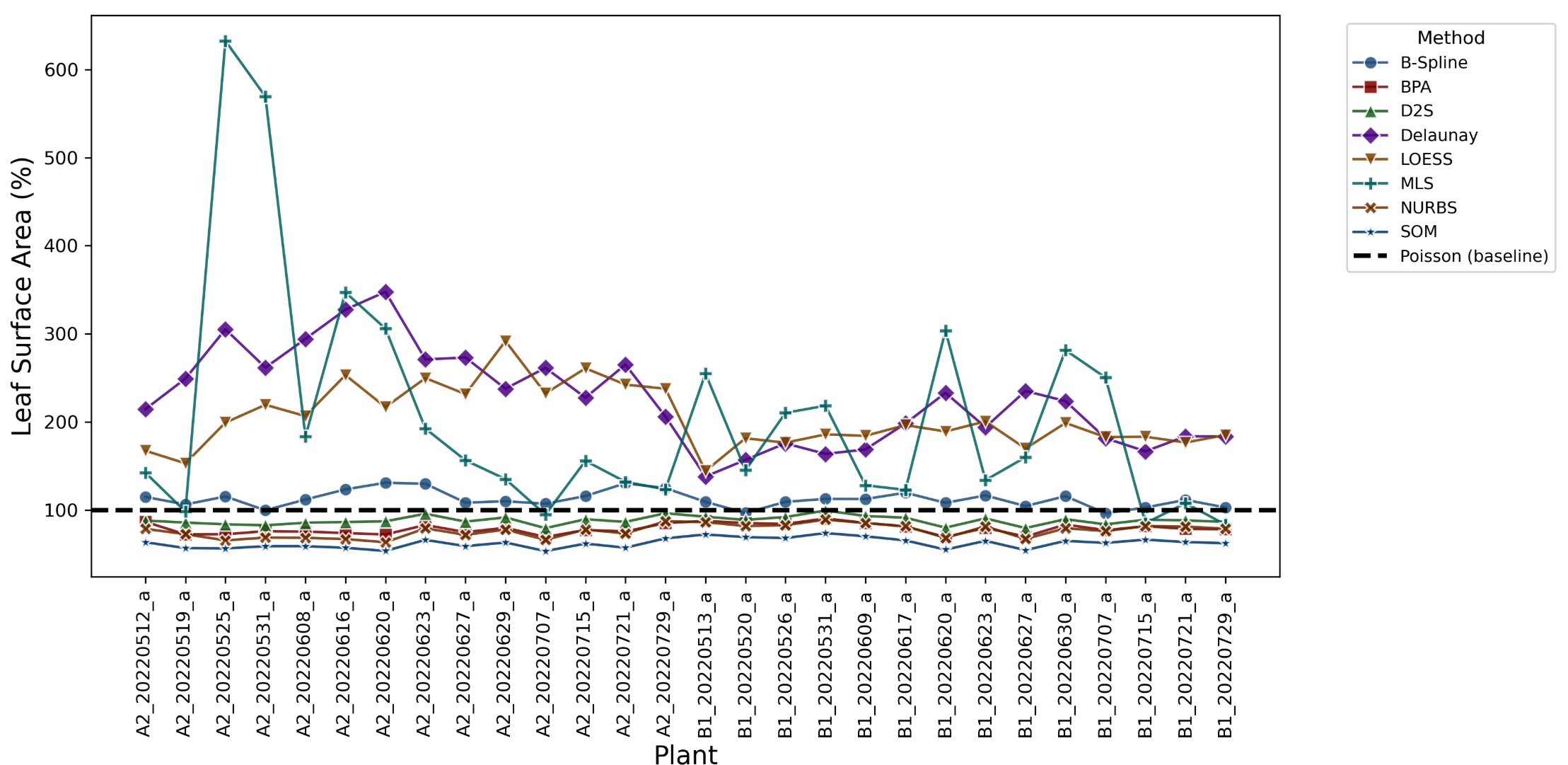}
    \caption{Mean leaf surface area (\%) for each plant computed by different methods in Sec.~\ref{sec:all_lsr_methods} on $\mathcal{D_{STRAW}}$ dataset  using Poisson as benchmark } 
    \label{fig:mean_leaf_area_per_plant}
\end{figure*}

\subsection{Implementation Details}
We implement all nine surface reconstruction methods in Python. We first denoise each point cloud using statistical outlier removal and rotate it into a PCA-aligned frame so that the leaf lies flat in the $uv-$plane \cite{santos20143d}. We implement B-spline interpolation and NURBS using \textit{geomdl} \cite{bingol2019geomdl} and, resampling the point cloud onto a $40\times40$ parametric grid before surface fitting. Since both methods operate on a rectangular parametric grid that extends beyond the actual leaf boundary, we fit a B-spline trim curve to restrict area integration within the leaf boundary \cite{flory2008constrained, santos20143d}.
For Delaunay 2.5D, MLS, and LOESS, we follow the approach described  in the literature \cite{zhu20183d}. We project the PCA-rotated point cloud to the $uv$-plane, apply each method independently, apply Delaunay triangulation to the resulting $uv$ point locations, and construct TriangleMesh using Open3D with the 3D coordinates. 
We implement BPA and Poisson surface reconstruction methods using Open3D and estimate surface normals. For BPA, the ball radius is set adaptively as a multiple of the mean nearest neighbor distance.  For Poisson, we use octree depth of 9. 
For the $D^2$-spline, we implement a close approximation to the method of \cite{kempthorne2015surface}. We use Scipy's thin-plate-spline basis function instead of Hsieh-Clough-Tocher finite element basis function and setting the $\alpha$ empirically rather than via GCV - both of which require a custom finite element solver beyond the scope of our comparative study. We avoid center-line parameterization for twisted wheat leaves but do PCA rotation, which is more appropriate for our study. 
Since none of the above-described reconstruction methods provide native GPU support, we run all nine implementations on CPU using the libraries described above, including our own SOM implementation with a $20 \times 20$ lattice (i.e, neuron grid).
 
 \begin{figure}[]
    \centering
    \includegraphics[width=0.99\linewidth]{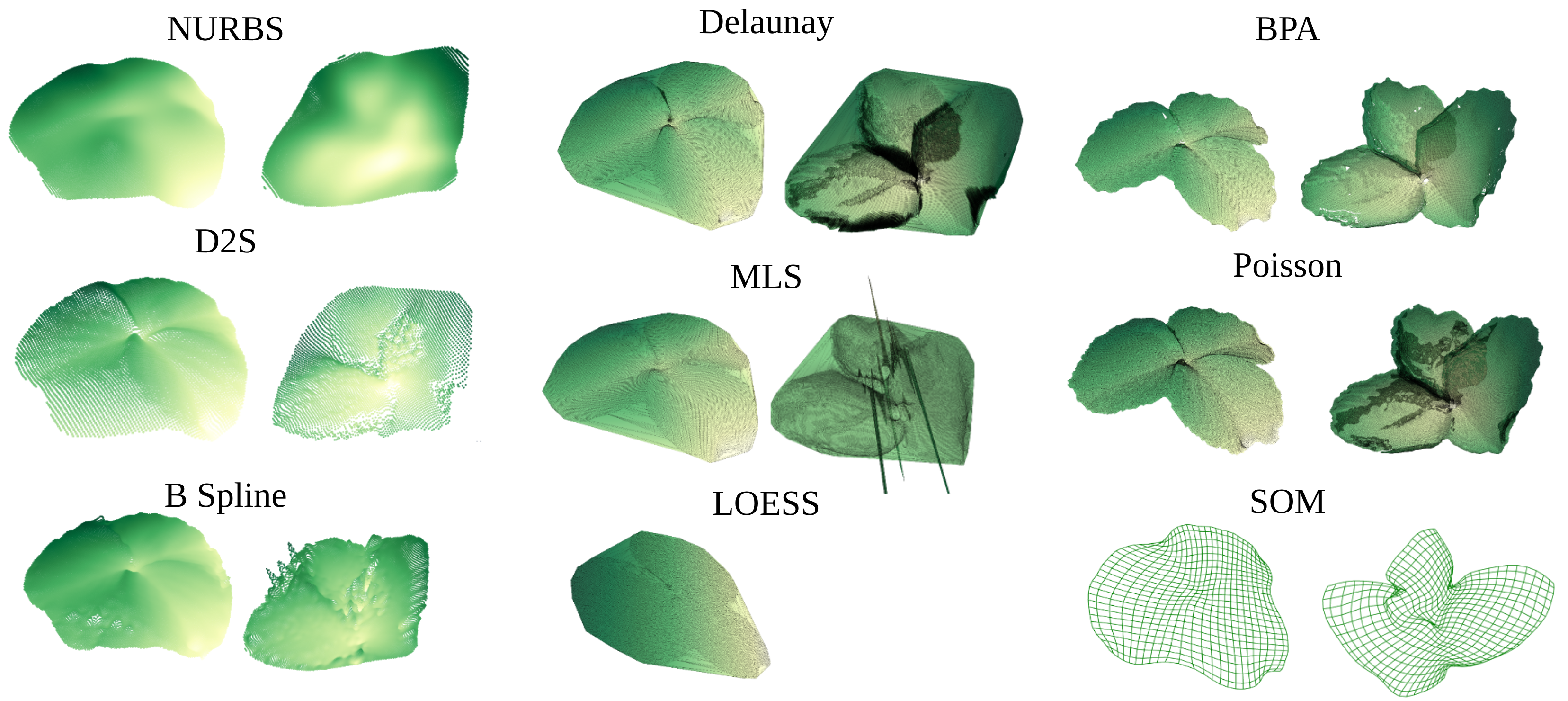}
    \caption{Reconstruction of Trifoliate Strawberry Leaf Clusters}
    \label{fig:strawberry_leaf_cluster}
\end{figure}

\begin{figure*}[!htbp]
    \centering
    \begin{subfigure}[b]{0.48\textwidth}
    \centering
\includegraphics[width=0.99\linewidth]{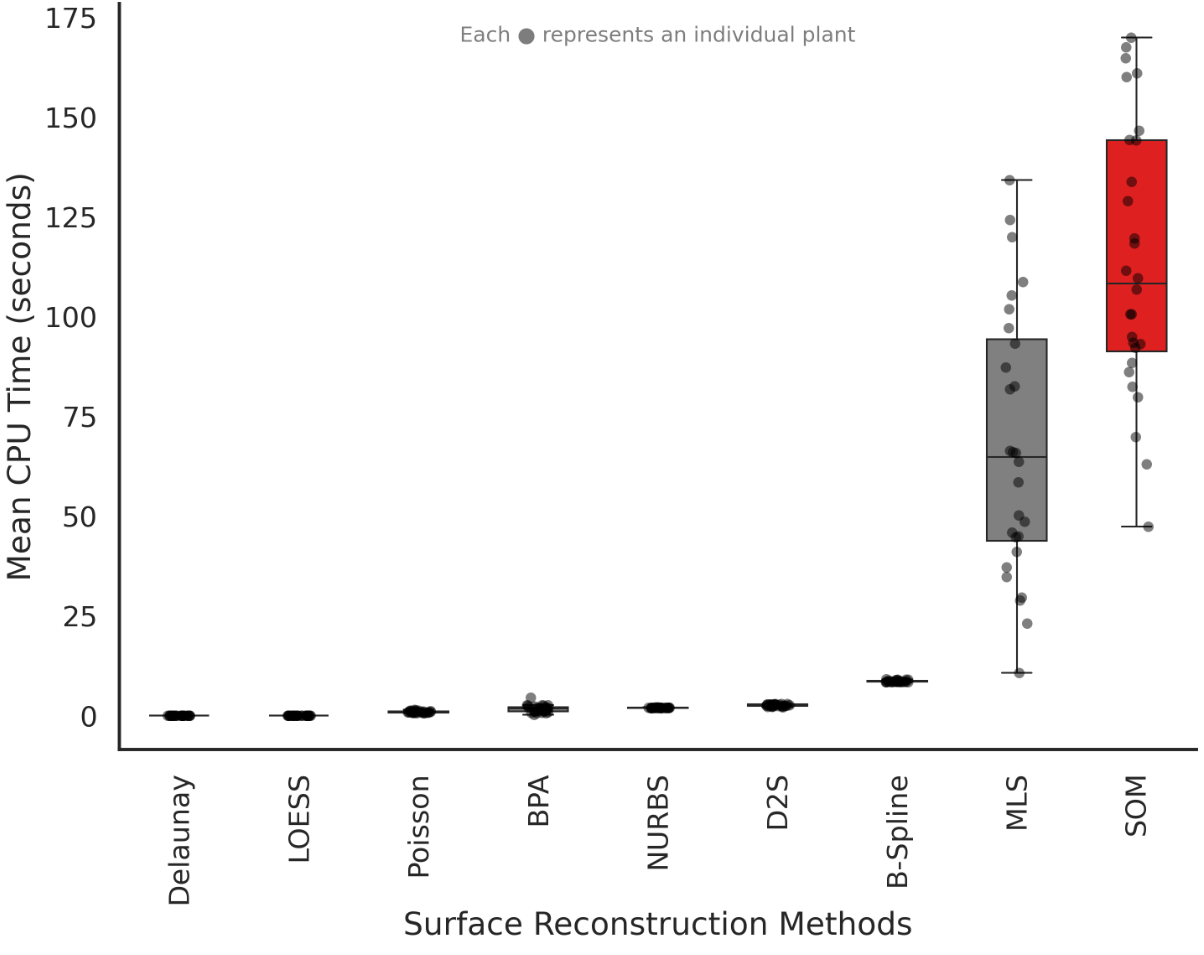}
    \caption{ Comparison on mean CPU time per plant across leaf surface reconstruction methods   }
    \label{fig:cpu_exec_per_plant}
    \end{subfigure}
    \hfill
    \begin{subfigure}[b]{0.48\linewidth}
    \centering
    \includegraphics[width=0.99\textwidth]{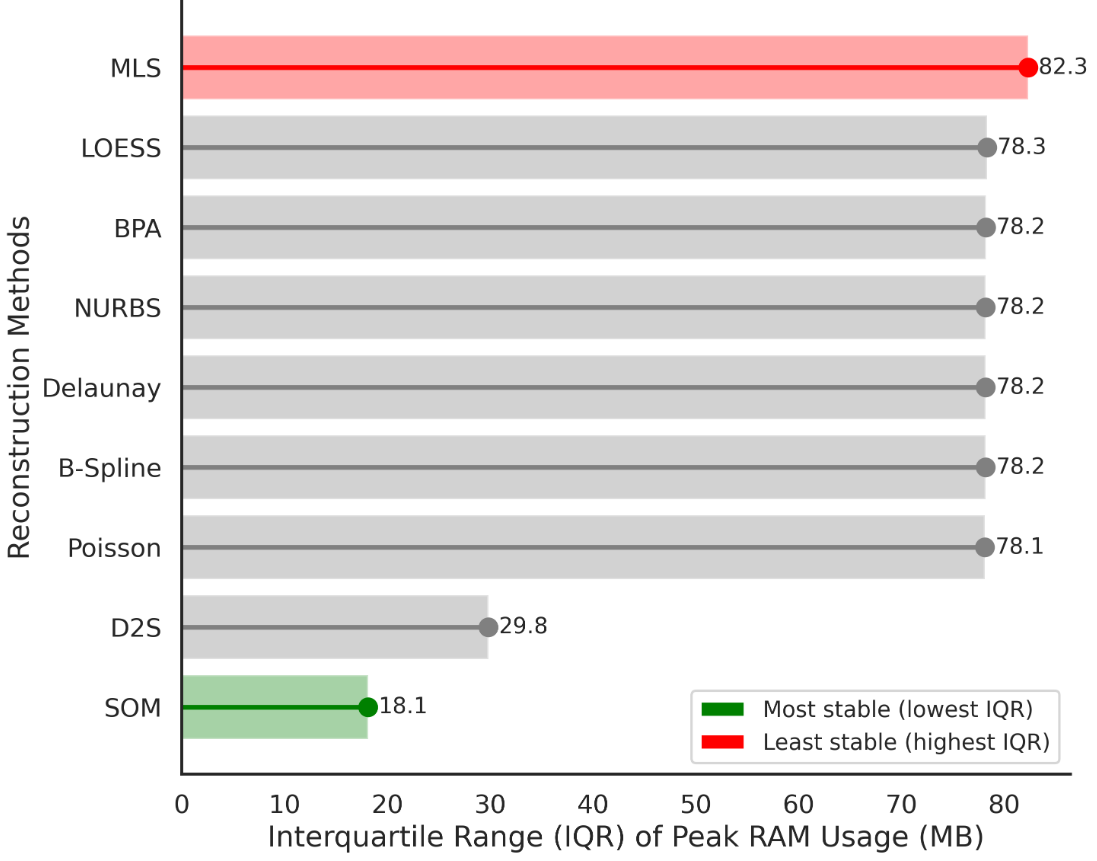}
    \caption{ Variability of peak RAM usage across leaf surface reconstruction methods }
    \label{fig:ram_peak_usg}
    \end{subfigure}
    \label{fig:combined_mean_cpu_peak_ram}
    \caption{Comparison of Leaf Surface Reconstruction Methods on the  $\mathcal{D_{STRAW}}$ Dataset  (a) Mean CPU Time per Plant (b) Variability of Peak RAM Usage (IQR)}
\end{figure*}

\subsection{Leaf Surface Reconstruction Results}
In Figure~\ref{fig:leaflet_lsr_results},  we show the leaf surface reconstruction results of a randomly selected strawberry leaflet. 
All parametric methods (B-Spline, NURBS, and $D^2$-Spline), all of them generated smoother surfaces. However, $D^2$-spline resulted much leaf surface, specifically leaf-tip. B-spline produces spikes for the leaf tip, which is known and expected for this method. Since NURBS use rational weights, it resulted comparatively better than B-Spline. 
The three methods Delaunay, MLS, and LOESS follow the same pipeline but we see smoother surfaces for MLS and LOESS as they applied polynomial interpolation. For MLS and LOESS we can see more triangles than Delaunay. 
However, LOESS considered the leaf tip points as noise and also much smoother leaf edges. Both Delaunay and MLS connected the triangulated points incorrectly which produce larger surface area error. 
Among the mesh-based and across all methods, the Poisson reconstructs the most accurate leaf surface. The BPA does mostly accurate but for the leaf tip the ball-radius was not appropriate to reconstruct that part sufficiently accurate. 
The learning based SOM also better captures the leaf topology. This unsupervised method can produce much better results with increased lattice (i.e, grid) and epochs. Besides, this method is robust to leaf missing points and holes. 

For all 893 leaves from $\mathcal{D_{STRAW}}$, we report the reconstructed surface area and CPU time in Table~\ref{tab:leaflet_results}. The Poisson provides most accurate surface area but for increased points it consumes more CPU memory  and execution time. 
Considering Poisson as the benchmark method, we present the percentage deviation (overestimation and underestimation) in reconstructed surface area (per plant) of each methods in Figure~\ref{fig:mean_leaf_area_per_plant}.
Delaunay, LOESS, and MLS exhibits notable overestimation, with MLS showing the largest spikes. B-splines shows very close agreement with overestimation. 
Other methods show close agreement with underestimation.

In Figure~\ref{fig:cpu_exec_per_plant}, we see SOM requires more CPU time for it's iterative learning process. However, this method consumes the lowest peak RAM consumption of $ 79.3 ~\mathrm{MB}$ ($3 \times$ lower than D2S). Our obtained ranking is: 
\(
\text{SOM} \ll \text{LOESS} < \text{Delaunay} < \text{NURBS} < \text{B-Spline} < \text{BPA} < \text{Poisson} \ll \text{MLS} \lll \text{D2S} 
\).
Moreover, we also show in  Figure \ref{fig:ram_peak_usg} that SOM exhibits the smallest IQR, indicating that variability in leaf point clouds has a limited effect on its peak RAM consumption. 

In Table~\ref{tbl:c3d_p4d}, we report the leaf surface reconstruction results for eight leaves from Crops3D dataset, these leaf point clouds were constructed from LiDAR, MVS, and structured-light in real-world agricultural environments. So, the number of points in these leaf point clouds is lower than other two datasets, contain noisy or missing points. 
Under these challenging conditions, SOM demonstrates more consistent and robust performance compared to other methods, as evidenced in Table~\ref{tbl:c3d_p4d}. 
However, we observe a tendency of SOM to  underestimate leaf surface area. This limitations can be addressed by increasing the number of SOM training epochs, which improves convergence and enhances surface reconstruction quality, particularly for high-resolution point cloud.
Even with fewer iterations, SOM is able to capture the overall leaf topology across most leaves, highlighting its robustness, even though with reduced geometric precision.

For broad and flat leaves, specifically cabbage and tomato, all reconstruction methods performed well, produced visually coherent and geometrically consistent surfaces. 
Low-resolution leaf point clouds with discontinued, noisy, or missing points, BPA, despite its fragmented and incomplete surface coverage, consistently captured the most geometrically accurate leaf shape after SOM. 
In these cases, Poisson produced incorrect shapes with extreme overestimation of leaf surfaces (for rapeseed, rice, and wheat). 
However, for maize ($\mathcal{D}_{C3D}$)  with 43K points, Poisson produces the best surface with fine geometric details despite that there were significant missing or discontinued points. 
MLS failed to reconstruct meaningful surfaces for three of the leaves (maize, potato, wheat), extremely overestimating surfaces, and using most of the available memory during computation. In contrast, LOESS with tri-cube-weight handled discontinuity or missing data points more effectively. 

For robotic platforms that use single board computers (SBCs) such as Raspberry Pi, we find that SOM remains a practical option for leaf surface reconstruction despite its longer processing time, whereas memory-intensive methods become impractical under such constrained computational conditions. 

For qualitative analysis, we present Figure~\ref{fig:strawberry_leaf_cluster} to compare nine surface reconstruction methods applied to a point cloud of trifoliate leaf clusters from LAST-STRAW.  
We observe that spline-based methods (first column of Figure~\ref{fig:strawberry_leaf_cluster}) overestimate the surface and fail to capture the concave regions between the lobes, which results in an overly smooth representation that does not reflect the true geometry. 
In the second column of the Figure~\ref{fig:strawberry_leaf_cluster}, we find that triangulation and fitting-based methods incorrectly connect points that are geometrically far apart or belong to different leaflet surfaces, creating triangles that fill in regions that should remain open or concave. 
We observe that BPA, shown in the top right of Figure~\ref{fig:strawberry_leaf_cluster}, reconstructs reasonably accurate with a small hole. We find that Poisson reconstruction performs the best overall, as it produces a clean and complete surface that closely follows the natural curvature of the leaf cluster. Lastly, we see that SOM, shown in the bottom right of Figure~\ref{fig:strawberry_leaf_cluster}, captures the overall shape adequately but lacks the fine surface detail along edges.

   
\begin{table*} 

\caption{Plant-level leaf (leaflet) surface reconstruction results on 
$\mathcal{D_{STRAW}}$ dataset }\label{tab:leaflet_results}
\centering 
\setlength{\tabcolsep}{2pt} 
\renewcommand{\arraystretch}{2.0}
\footnotesize 
\begin{tabularx}{\textwidth}
{@{}C|CC|CC|CC|CC|CC|CC|CC|CC|CC@{}}
\toprule
   
      & \multicolumn{2}{c}{Poisson} & \multicolumn{2}{c}{BPA} & \multicolumn{2}{c}{B-spline} & \multicolumn{2}{c}{NURBS} & \multicolumn{2}{c}{Delaunay} & \multicolumn{2}{c}{MLS} & \multicolumn{2}{c}{LOESS} & \multicolumn{2}{c}{SOM}  & \multicolumn{2}{c}{$D^2$-spline}  \\ 
      
  Plant ID & Area & CPU & Area & CPU& Area & CPU& Area & CPU& Area & CPU& Area & CPU& Area & CPU& Area & CPU & Area & CPU  \\
    
\midrule
\midrule

A.512a & 2461.06 & 0.73 & 2129.99 & 0.77 & 2827.26 & 8.55 & 1934.06 & 2.01 & 5277.37 & 0.03 & 3502.87 & 29.69 & 4116.21 & 0.04 & 1560.40 & 69.90 & 2165.88 & 2.27 \\
A.519a & 3821.52 & 1.03 & 2762.05 & 2.10 & 4059.52 & 8.51 & 2771.82 & 1.95 & 9520.69 & 0.04 & 3753.34 & 63.72 & 5845.22 & 0.06 & 2168.87 & 111.62 & 3276.19 & 2.77 \\
A.525a & 3922.36 & 0.90 & 2842.58 & 1.91 & 4524.57 & 8.52 & 2568.91 & 2.02 & 11967.65 & 0.04 & 43986.13 & 66.07 & 7813.24 & 0.06 & 2210.73 & 169.98 & 3284.84 & 2.50 \\
A.531a & 4288.99 & 1.11 & 3257.71 & 2.35 & 4274.05 & 8.58 & 2945.52 & 2.04 & 11222.42 & 0.05 & 24417.42 & 87.32 & 9411.51 & 0.07 & 2526.22 & 129.02 & 3543.32 & 2.63 \\
A.608a & 5435.24 & 1.26 & 4101.89 & 2.10 & 6080.32 & 8.86 & 3720.21 & 2.09 & 15991.70 & 0.06 & 32390.72 & 101.93 & 11226.12 & 0.08 & 3202.18 & 146.63 & 4649.28 & 2.95 \\
A.616a & 5378.83 & 1.19 & 3971.17 & 2.13 & 6635.48 & 8.61 & 3602.31 & 2.05 & 17612.71 & 0.06 & 18670.73 & 108.77 & 13615.28 & 0.08 & 3070.87 & 161.02 & 4644.61 & 2.97 \\
A.620a & 6190.73 & 1.50 & 4468.68 & 4.60 & 8104.98 & 9.21 & 3923.15 & 2.10 & 21521.95 & 0.07 & 18936.12 & 134.22 & 13418.79 & 0.09 & 3314.71 & 164.79 & 5393.46 & 2.92 \\
A.623a & 4882.19 & 1.27 & 4051.48 & 2.06 & 6334.37 & 8.93 & 3864.36 & 2.07 & 13224.15 & 0.05 & 36812.88 & 93.35 & 12191.04 & 0.08 & 3226.56 & 144.38 & 4673.72 & 2.91 \\
A.627a & 5643.79 & 1.27 & 4225.41 & 2.65 & 6114.50 & 8.54 & 4052.87 & 2.08 & 15415.39 & 0.06 & 8825.97 & 120.00 & 13068.55 & 0.09 & 3327.99 & 167.56 & 4899.22 & 2.97 \\
A.629a & 5106.95 & 1.31 & 4075.59 & 2.59 & 5613.75 & 8.72 & 3968.38 & 2.13 & 12137.73 & 0.06 & 6898.72 & 105.37 & 14900.27 & 0.08 & 3222.37 & 144.26 & 4676.81 & 2.87 \\
A.707a & 5657.77 & 1.39 & 3923.68 & 2.68 & 6058.78 & 9.11 & 3745.51 & 2.09 & 14786.38 & 0.06 & 5358.13 & 124.28 & 13145.06 & 0.09 & 3018.20 & 160.10 & 4490.70 & 2.92 \\
A.715a & 4880.36 & 1.12 & 3775.81 & 1.91 & 5653.90 & 8.54 & 3796.57 & 2.05 & 11104.88 & 0.05 & 7602.56 & 82.56 & 12728.74 & 0.07 & 3018.60 & 133.90 & 4368.80 & 2.94 \\
A.721a & 4655.13 & 1.04 & 3523.03 & 2.61 & 6068.20 & 8.57 & 3417.31 & 2.01 & 12339.09 & 0.04 & 6140.30 & 81.91 & 11282.67 & 0.06 & 2663.06 & 119.63 & 4026.05 & 2.46 \\
A.729a & 3385.25 & 0.97 & 2881.13 & 1.33 & 4224.97 & 8.61 & 2950.89 & 1.96 & 6976.35 & 0.04 & 4177.62 & 58.58 & 8045.13 & 0.05 & 2295.17 & 100.62 & 3259.85 & 2.40 \\
\hline 
B.513a & 1674.26 & 0.72 & 1466.33 & 0.32 & 1828.90 & 9.01 & 1443.91 & 1.99 & 2309.99 & 0.02 & 4269.17 & 10.75 & 2415.61 & 0.03 & 1209.22 & 47.40 & 1546.64 & 2.15 \\
B.520a & 2277.36 & 0.71 & 1940.69 & 0.73 & 2200.55 & 8.51 & 1862.20 & 2.01 & 3572.24 & 0.03 & 3306.75 & 23.20 & 4132.28 & 0.04 & 1576.27 & 63.02 & 2021.21 & 2.37 \\
B.526a & 2736.64 & 0.73 & 2303.64 & 0.88 & 2989.23 & 8.50 & 2264.43 & 2.01 & 4806.75 & 0.03 & 5754.96 & 28.96 & 4826.97 & 0.04 & 1865.47 & 79.92 & 2517.95 & 2.55 \\
B.531a & 3185.20 & 0.77 & 2878.64 & 0.64 & 3587.72 & 8.46 & 2837.18 & 1.98 & 5211.17 & 0.03 & 6955.59 & 37.24 & 5918.82 & 0.05 & 2351.22 & 93.18 & 3166.12 & 2.82 \\
B.609a & 2989.16 & 0.77 & 2549.95 & 0.78 & 3361.02 & 8.42 & 2546.00 & 1.93 & 5045.66 & 0.03 & 3827.70 & 34.78 & 5506.43 & 0.05 & 2096.22 & 86.13 & 2785.90 & 2.75 \\
B.617a & 3368.48 & 0.90 & 2736.92 & 1.35 & 4029.50 & 8.52 & 2746.96 & 2.02 & 6676.29 & 0.04 & 4137.31 & 50.28 & 6613.92 & 0.05 & 2205.60 & 93.48 & 3073.58 & 2.72 \\
B.620a & 4106.76 & 1.00 & 2846.40 & 2.05 & 4439.27 & 8.58 & 2809.48 & 2.07 & 9567.87 & 0.04 & 12465.27 & 97.22 & 7767.66 & 0.06 & 2259.09 & 109.71 & 3283.51 & 2.74 \\
B.623a & 3221.93 & 0.86 & 2568.10 & 1.09 & 3754.15 & 8.58 & 2624.91 & 2.02 & 6248.32 & 0.04 & 4312.61 & 45.90 & 6462.07 & 0.05 & 2093.93 & 92.29 & 2914.68 & 2.54 \\
B.627a & 3871.59 & 1.09 & 2701.88 & 1.70 & 4041.71 & 9.13 & 2598.94 & 2.03 & 9104.24 & 0.04 & 6189.23 & 66.48 & 6575.18 & 0.06 & 2106.66 & 106.85 & 3078.48 & 2.52 \\
B.630a & 3862.87 & 0.96 & 3224.23 & 2.73 & 4468.68 & 8.49 & 3063.15 & 1.99 & 8636.94 & 0.04 & 10872.31 & 65.92 & 7679.56 & 0.06 & 2509.59 & 118.37 & 3458.81 & 2.77 \\
B.707a & 3331.74 & 0.85 & 2568.11 & 1.09 & 3207.20 & 8.55 & 2535.07 & 1.97 & 6048.44 & 0.04 & 8338.08 & 48.69 & 6087.21 & 0.05 & 2086.43 & 94.99 & 2789.78 & 2.49 \\
B.715a & 3224.92 & 0.93 & 2630.94 & 1.12 & 3316.47 & 9.11 & 2637.86 & 2.01 & 5373.12 & 0.04 & 2741.20 & 44.79 & 5907.51 & 0.05 & 2139.84 & 88.43 & 2868.32 & 2.50 \\
B.721a & 3341.18 & 0.85 & 2625.58 & 1.34 & 3724.23 & 8.56 & 2710.03 & 1.99 & 6134.84 & 0.03 & 3603.73 & 41.15 & 5899.40 & 0.05 & 2124.23 & 82.50 & 2947.35 & 2.49 \\
B.729a & 3394.78 & 0.79 & 2651.33 & 1.50 & 3490.78 & 8.53 & 2674.59 & 1.98 & 6231.56 & 0.04 & 2828.10 & 45.00 & 6283.09 & 0.05 & 2116.35 & 100.67 & 2936.13 & 2.30 \\
\bottomrule
\bottomrule
\multicolumn{19}{c}{}   \\ 

\multicolumn{19}{c}{Here, ``Area'' indicates mean total leaf surface area per plant, while ``CPU'' denotes mean computation time per plant.}
\end{tabularx}

\end{table*}

\section{Robotic Perception Challenges for  Leaf Surface Reconstruction}
\label{sec:4_robotic_percept}
In this section, we address the challenges of 3D perception in real-world agricultural environments, which directly affect downstream leaf surface reconstruction tasks.

\subsection{3D Sensing in Indoor and  Complex Agricultural Environment}

\cite{zermas2017estimating} capture multiview camera images  of corn plants, use VisualSFM for 3D point cloud reconstruction, apply Euclidean clustering to segment leaf point clouds,  and estimate leaf surface area using Self-Organizing map \cite{kohonen1997exploration}. 
\cite{wu2022miniaturized} proposed a platform that reconstructs individual plants using multiview stereo 3D reconstruction. This platform requires the plant to be placed within it, where the fixed camera arms rotate and capture multi-view images.  They also minimize the light and wind effects to obtain high-quality images for 3D reconstruction. Then, they estimate phenotypic traits such as plant height estimation and leaf area estimation.  
\cite{wei2024fast} use a turntable setup, where they place a plant and rotate the table uniformly. To reduce the impact of natural light on the sensor data, they cover the interior chamber with a black background, LED light strips, and a diffused fabric to provide uniform and stable lighting. 

Moreover, in our literature review, we found several agricultural robotics works that use ground robots equipped with a robotic arm (i.e., mobile manipulator) to solve the next-best-view (NBV) planning problem for plants or plant parts reconstruction
\cite{menon2023nbv, burusa2024gradient, burusa2024attention, isaacjose2025iros, ci2025ssl}.
\cite{roy2017active} proposed an algorithm for apple counting in orchards from geometric information of 3D point clouds.
However, most 3D reconstruction works are conducted in controlled environments and often overlook environmental factors such as wind and rain, which can cause leaf movements \cite{sampaio20213d}.

\subsection{Real-time Leaf Surface Reconstruction for Ag-robotics}
In this comparative study, we highlight a core trade-off between reconstruction accuracy and practical deployability.  
Spline-based methods produce smooth surface but rely on computationally expensive iterative fitting and parameterization, limiting real-time use. 
Mesh-based approaches are more flexible, but suffer from noise sensitivity, high local memory usage, or incomplete reconstruction in sparse regions. 
The SOM offers a structured and smooth surface representation through  competitive learning, yet requires lengthy unsupervised training and lacks adaptability in resolution. So, there is a need for reconstruction method for real-time deployment.


\section{Open Challenges and Research Directions}
\label{sec:5_open_challenge}
Despite the progress demonstrated in this study, several important challenges remain that require further investigation on leaf surface reconstruction.

All surface reconstruction methods were  evaluated on relatively clean point clouds acquired under controlled conditions \cite{uol2024laststraw,schunck2021pheno4d}. 
In real-world agricultural field, data quality often  degraded by occlusion from overlapping leaves, wind-induced motion, varying illumination, and sensor specific noise patterns associated with LiDAR, structured light, or stereo imaging systems \cite{li2025applications}.
Robot motion during scanning introduces additional registration errors that compound with sensor noise. 
A particularly challenging scenario arises in UAV deployments, where downwash effect cause leaves to oscillate and deform dynamically while scanning. 
Developing reconstruction methods that are robust to such degradations, or that explicitly account for sensing uncertainty, remains a key challenge for Ag-robotics applications. 
A further challenge lies in view planning, as limited viewpoints often lead to incomplete point cloud observations, thereby preventing accurate and complete reconstruction of leaf surfaces \cite{burusa2024attention}. 

A fundamental challenge that underlies all reconstruction methods is the assumption that individual leaves have already been isolated from the full plant point cloud. In practice, leaf segmentation directly conditions leaf surface reconstruction quality. More specifically, the errors at the segmentation stage propagate into the leaf surface reconstruction, which affects phenotyping tasks such leaf area estimation. \cite{zermas2017estimating} demonstrate this dependency in their work on corn plants, where their segmentation approach using Euclidean Clustering and then Skeleton-Kalman-Filer (SKF) tends to over-segment individual leaves into many parts. As a result, the segmentation errors systematically introduce underestimation of leaf surface area. 
Later, \cite{masuda2021leaf} used PointNet++ for semantic segmentation of tomato leaves and estimate leaf area from number of leaf points around the stem. Their method without reconstructing the actual leaf surface geometry produced around $20\%$ error, highlighting the need for accurate leaf surface reconstruction. 
Therefore, developing jointly learned segmentation and reconstruction pipelines remains a critical open direction for accurate whole-plant leaf area estimation.

SOM-based approximations of leaf surfaces   may not fully capture the true underlying geometry and micro-details such as leaf veins, which can affect other phenotypic tasks such as water consumption \cite{zhu2024crops3d, hui2026leaflods}.
Higher-quality ground truth is difficult to obtain at scale from physical measurements alone. A promising direction is the use of simulation pipelines that generates photorealistic synthetic plants  with exact geometric ground truth. \cite{dangplant2sim3d} propose a pipeline for reconstructing 3D plant meshes from monocular video using Neural Radiance Fields (NeRF), providing a route to generate accurate surface meshes that could support training models. Such simulation-to-real pipelines offer a promising avenue, although their transferability to real sensor data remains an open question. 
Each reconstruction methods performance on edge devices such as NVIDIA Jetson or Raspberry Pi has not yet been systematically evaluated. Such evaluation is necessary before field deployment.

\section{Conclusion}
\label{sec:6_conclusion}
Accurate leaf surface area estimation from 3D point cloud data is a fundamental requirement for precision agriculture and agricultural robotics, yet no systematic comparison of reconstruction methods exists to guide method selection for deployment.
In this work, we evaluate nine leaf surface reconstruction methods across common datasets and implementation framework, providing comprehensive analysis.
Poisson reconstruction provided the most accurate  surface for clean and high-resolution point cloud but struggled for low-resolution and noisy data. Parametric methods such as B-Spline, NURBS, $D^2$-Spline produced smooth surfaces but failed to capture geometry for complex leaf point cloud data. BPA performed well for noisy data but resulted fragmented surfaces due to inappropriate ball radius.  
SOM emerged as a strong alternative, offering robust topology capture with the lowest peak RAM consumption, making it well-suited for deployment on resource constrained robotic platforms.

Our analysis reveals that parametric surface methods produce smooth surfaces but iterative fitting limits real-time use. Implicit surface methods provide the most accurate surfaces, but they are computationally expensive for high-resolution point cloud data.
No existing method simultaneously satisfies the accuracy, speed, and deployment constraints required for on-device inference on agricultural robots. 
This work contributes a systematic evaluation framework for leaf surface reconstruction   for real-time deployment. 
Our future work will address generalization, robustness to occlusion and sensor noise in agricultural field conditions.

\printcredits

\bibliographystyle{cas-model2-names}



\end{document}